\documentclass{article}
\PassOptionsToPackage{numbers, compress}{natbib}

\usepackage[final]{neurips_2025}

\usepackage[utf8]{inputenc} 
\usepackage[T1]{fontenc}    
\usepackage{hyperref}       
\usepackage{url}            
\usepackage{booktabs}       
\usepackage{amsfonts}       
\usepackage{nicefrac}       
\usepackage{microtype}      
\usepackage{xcolor}         
\usepackage{graphicx} 
\usepackage{multirow}
\usepackage{amsmath}
\usepackage{subcaption}
\usepackage{enumitem}
\newcommand{\bigO}{\mathcal{O}}

\title{
\begin{minipage}[c]{0.1\textwidth} 
    \includegraphics[width=\linewidth]{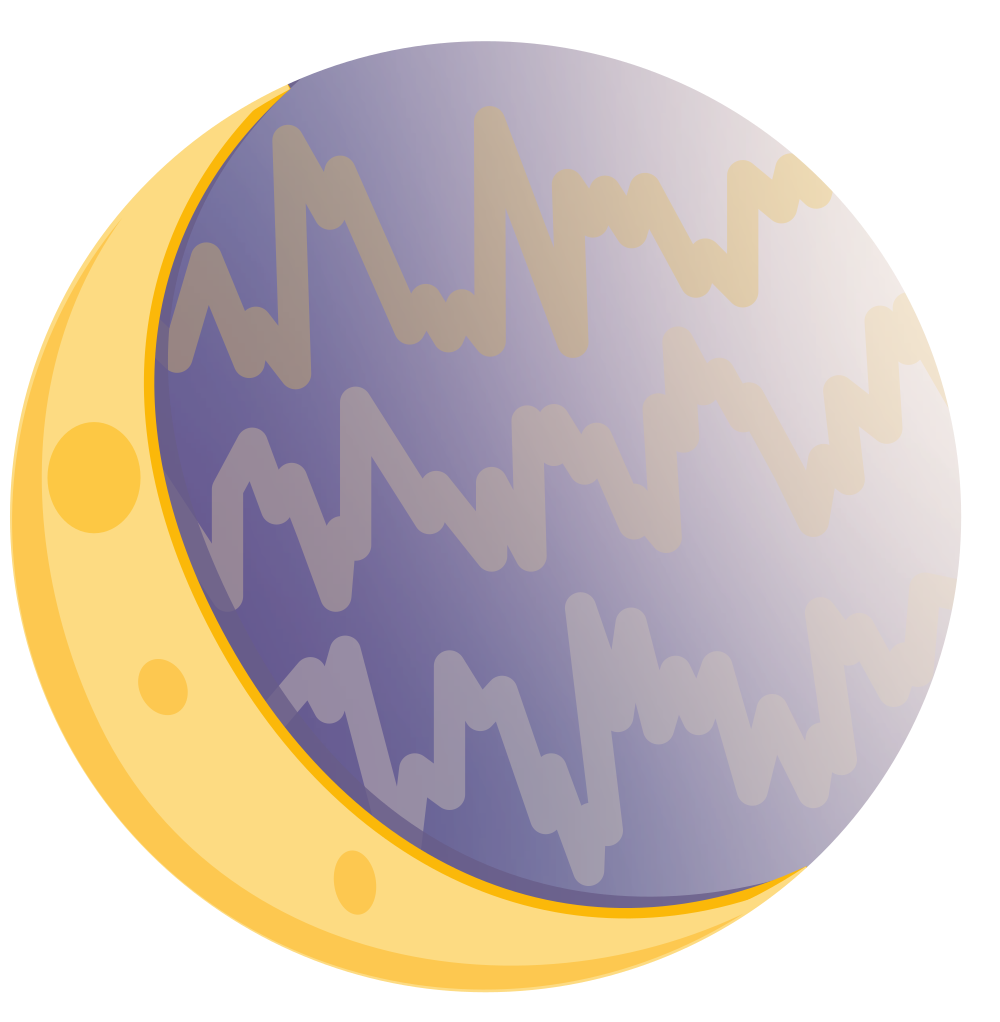}
\end{minipage}
\begin{minipage}[c]{0.82\textwidth}
    \raggedright
    \Large
    LUNA: Efficient and Topology-Agnostic Foundation Model for EEG Signal Analysis 
\end{minipage}
}

\author{%
  Berkay Döner\textsuperscript{1}
  \quad
  Thorir Mar Ingolfsson\textsuperscript{1}\thanks{Correspondence to \texttt{\{thoriri,yawli\}@iis.ee.ethz.ch}.}
  \quad
  Luca Benini\textsuperscript{1} \quad
  Yawei Li\textsuperscript{1}\\
  \textsuperscript{1}Integrated Systems Laboratory, ETH Z{\"u}rich, Switzerland}

\begin{document}

\maketitle

\begin{abstract}
  Electroencephalography (EEG) offers a non-invasive lens into human brain activity, but building large‐scale models is hampered by \emph{topological heterogeneity}: each public EEG data defines its own electrode layout, limiting generalization. We introduce \textbf{LUNA} (\textbf{L}atent \textbf{U}nified \textbf{N}etwork \textbf{A}rchitecture), a self-supervised foundation model that reconciles disparate electrode geometries while scaling linearly---not quadratically---with channel count. LUNA compresses multi-channel EEG into a fixed-size, topology-agnostic latent space via \emph{learned queries} and cross-attention. Downstream transformer blocks then operate exclusively on this latent representation using patch-wise temporal self-attention, decoupling computation from electrode count. Pre-trained on TUEG and Siena ($>$ 21,000 hours of raw EEG across diverse montages) using a masked-patch reconstruction objective, LUNA transfers effectively to four downstream tasks: abnormality detection, artifact rejection, slowing classification, and emotion recognition. It demonstrates highly competitive performance across several benchmarks, achieving state-of-the-art results on TUAR and TUSL, e.g., \textbf{0.921 AUROC} on TUAR, while reducing FLOPs by \textbf{300$\times$} and trimming GPU memory use by up to \textbf{10$\times$}. Critically,  these gains are consistent across all evaluated electrode configurations. Code is available at \url{https://github.com/pulp-bio/biofoundation}
\end{abstract}

\section{Introduction}
\label{sec:introduction}
Electroencephalography (EEG) provides deep insight into brain activity without requiring invasive procedures, and plays a crucial role in clinical diagnostics, cognitive neuroscience, and human-computer interaction. In recent years, deep neural networks have significantly advanced EEG analysis, shifting from handcrafted pipelines to end-to-end learning systems~\cite{craik2019deep}. Transformer-based models now rival traditional signal processing techniques by jointly modelling long-range temporal dynamics and cross-channel correlations~\cite{song2022eeg,wen2022transformers}. 

Despite this progress, \emph{a fundamental bottleneck remains}: EEG corpora exhibit significant \emph{topological heterogeneity}. Electrode count and placement vary widely across public and private datasets, making it difficult to transfer models across montages. This limitation manifests in pronounced performance degradation during cross-dataset evaluation. For example, motor-imagery decoders lose up to 14 percentage points (pp) in accuracy when transferring from PhysioNet to KU datasets~\cite{xu2020cross}, while state-of-the-art emotion-recognition models such as BIOT and MMM exhibit 13–15 pp drops between SEED and DEAP montages~\cite{yang2023biot,yi2023learning}. Similarly, patient-to-patient transfer in stereotactic EEG (sEEG) remains an unsolved challenge, with naive models performing near chance without explicit spatial encoding~\cite{mentzelopoulos2024neural}. 

Existing approaches offer limited solutions to this problem. Some train bespoke models for each montage, while others retain only shared electrodes---discarding up to \textbf{80\%} of available data~\cite{lin2023eeg}. More general approaches that flatten channels and time into long sequences incur quadratic self-attention complexity, \(\bigO\!\bigl((S \cdot C)^{2}\bigr)\) where $S$ is the number of time segments and $C$ is the number of electrodes (channels), rapidly exhausting memory on dense caps~\cite{yang2023biot}. These challenges underscore the need for a \textbf{single, montage-agnostic architecture that scales efficiently with electrode count}.

\textbf{LUNA} (\textbf{L}atent \textbf{U}nified \textbf{N}etwork \textbf{A}rchitecture) directly addresses this gap. Our key innovation is a topology-invariant encoder that maps arbitrary electrode layouts into a fixed latent space via learned queries and cross-attention. Temporal self-attention layers then operate exclusively on this latent space, decoupling computational cost from the number of electrodes. We pre-train LUNA using a masked-patch reconstruction objective on \textsc{TUEG} \cite{obeid2016temple} and \textsc{Siena} \cite{detti2020siena} (over \textit{21,000} hours of raw EEG data), and fine-tune on four downstream benchmarks spanning abnormality and artifact detection, slowing classification, and emotion recognition.

The key contributions of this work are the following:
\begin{itemize}[leftmargin=*]
        \item \textbf{Topology-invariant encoder.} A learnt query / cross-attention module that projects arbitrary-sized channel sets into a fixed latent space.
        \item \textbf{Linear-in-channels complexity.} Patch-wise temporal attention that decouples FLOPs and memory from electrode count.
        \item \textbf{State-of-the-art accuracy-efficiency trade-off.} LUNA achieves strong results across a range of EEG benchmarks, demonstrating significant capabilities with balanced accuracies of \textbf{81.57\%} on TUAB and \textbf{39.18\%} on SEED-V \cite{liu2021comparing}, and AUROC scores of \textbf{0.921} on TUAR and \textbf{0.802} on TUSL, while reducing FLOPs by \textbf{300$\times$} and GPU memory footprint by up to \textbf{10$\times$} on high-density EEG recordings. Crucially, these gains hold across diverse electrode configurations, confirming LUNA's generalization capability.
\end{itemize}

\section{Related Work}
\label{sec:related_work}

To contextualize our contributions, this section discusses relevant state-of-the-art methodologies that we will compare against. We focus on advancements in self-supervised learning for time series, the emergence of foundation models for physiological signals, and existing approaches to managing variable input structures, especially concerning topological heterogeneity in the EEG domain and computational efficiency.

\subsection{Self-Supervised Learning Strategies in EEG}
Foundation models for EEG primarily rely on self-supervised learning (SSL) to leverage large unlabeled datasets. Masked signal modeling is a dominant paradigm. BENDR \citep{kostas2021bendr} pioneered this for EEG by adapting masked prediction concepts from speech, applying a contrastive objective to predict masked convolutional features. Subsequent models refined this: BrainBERT \citep{wang2023brainbert} performs masked prediction on channel-independent spectrograms for intracranial electroencephalography (iEEG); EEGFormer \citep{chen2024eegformer} and LaBraM \citep{jiang2024large} predict vector-quantized (VQ) representations of masked patches, learning discrete codebooks; CBraMod \citep{wang2024cbramod} directly reconstructs masked raw signal patches. LUNA employs a similar masked reconstruction objective but applies it after projecting channel information into a unified latent space, requiring the decoder to reconstruct channel-specific details from this compressed representation. 

\subsection{Modeling Spatial Structure and Topology Variation in EEG}
\label{subsec:related_work}
Capturing the spatial relationships between EEG channels is vital but complicated by varying electrode counts and layouts across datasets. Several strategies have been explored in the literature: \newline
\textbf{Channel Independence:} Early approaches and models like BrainBERT \citep{wang2023brainbert} and EEGFormer \citep{chen2024eegformer} process each channel's data independently before potentially combining them later. While inherently handling varying channel numbers, this neglects early modeling of cross-channel interactions. \newline
\textbf{Fixed-Topology Spatial Modeling:} Models like Brant \citep{zhang2023brant} use dedicated spatial encoders alongside temporal ones but assume a consistent channel configuration, limiting cross-dataset generalization. Graph Neural Networks (GNNs) \citep{tang2021self} explicitly model spatial relationships using a predefined adjacency graph, but require mechanisms to handle dynamically changing graph structures when topologies vary. LUNA avoids pre-defined graphs or fixed structures.  \newline
\textbf{Joint Spatio-Temporal Attention:} LaBraM \citep{jiang2024large} flattens channel and patch dimensions into one long sequence, allowing a standard Transformer to learn spatio-temporal dependencies simultaneously. However, this incurs $\bigO((S\cdot C)^2)$ complexity, scaling quadratically with both sequence length/patches (S) and channels (C). CBraMod \citep{wang2024cbramod} and CEReBrO \citep{dimofte2025cerebro} use alternating or parallel spatial and temporal attention mechanisms, reducing complexity to $\bigO(max(S^2, C^2))$ but still scaling quadratically with the dominant dimension. BIOT \citep{yang2023biot} uses linear attention after flattening, improving efficiency but potentially limiting modeling capacity. LUNA differs significantly by performing channel unification first before applying temporal attention with quadratic complexity only on the patch dimension and the much smaller latent dimension Q.  \newline
\textbf{Explicit Topology Mapping:} Some methods explicitly map varying topologies to a canonical representation. MMM \citep{yi2023learning} maps channels to predefined anatomical regions but relies on hand-engineered features (Differential Entropy) rather than raw signals. PopT \citep{chau2024population} aggregates pre-computed channel-independent temporal features using 3D electrode coordinates. While achieving topology invariance, these methods are not fully end-to-end or rely on external information (regions). LUNA learns an end-to-end mapping from raw signals using learned queries without requiring pre-defined structures.
\textbf{Differentiable Channel Reordering.}
Saeed et al.~\citep{saeed2021learning} also use learned attention to reorder/project heterogeneous channels into a fixed space, but under supervised training and without explicit 3D channel encodings. In contrast, LUNA is a self-supervised foundation model trained at scale, integrates explicit spatial (3D) information, and targets topology-agnostic \emph{and} compute-efficient transfer across datasets and tasks.
\subsection{Learned Queries and Efficient Attention for Set Abstraction}
LUNA's core mechanism for topology unification draws inspiration from architectures designed for permutation-invariant processing of set-structured data. Set Transformer \citep{lee2019set} introduced the concept of using a small set of learnable inducing points (queries) and an Induced Set Attention Block to summarize information from a larger input set via cross-attention, reducing the complexity from $\bigO(N^2)$ to $\bigO(M \cdot N)$. PerceiverIO \citep{jaegle2021perceiver} further developed this mechanism, demonstrating its power in creating a fixed-size latent bottleneck capable of handling diverse, variable-sized inputs across different modalities (images, text) and enabling flexible decoding via task-specific output queries.

LUNA adapts this principle specifically for EEG topology invariance. We treat the set of EEG channel features at a given time interval (patch) as the input set. By applying cross-attention between the channel features (as keys/values) and a small number (Q) of learned queries, LUNA projects the variable-channel input onto a fixed-size latent space ($\mathbb{R}^{Q \times E}$). This projection is permutation-invariant with respect to the input channels, thus achieving topology agnosticism. Furthermore, it improves computational efficiency, as the complexity of this step scales linearly with the number of channels. Where MMM relies on predefined regions (and hand-crafted features) and LaBraM/CBraMod rely on quadratic spatial attention after flattening space–time, LUNA first \emph{unifies} variable channel sets with learned queries (with explicit 3D encodings) and \emph{then} applies temporal attention on a fixed-size latent, yielding linear-in-$C$ unification and reduced temporal sequence length. This design specifically targets topology-agnostic scaling and inference efficiency within a self-supervised foundation-model framework, rather than combining prior components unchanged.

\section{Methodology}
\label{sec:method}

Developing generalizable foundation models for EEG is hindered by two primary obstacles: the \textbf{topological heterogeneity} of EEG montages (varying channel counts and layouts) and the \textbf{computational complexity} of attention mechanisms. Standard models struggle with diverse input channel configurations, limiting data aggregation and generalizability. Furthermore, transformer-based approaches often face prohibitive \(\bigO((C \cdot S)^2)\) or \(\bigO(\max(C^2, S^2))\), as discussed in the section \ref{subsec:related_work}, complexity when processing \(C\) channels and \(S\) temporal patches. This limits their applicability to high-density EEG or long recordings.

\begin{figure}[htpb!]
    \centering
    \includegraphics[width=\linewidth]{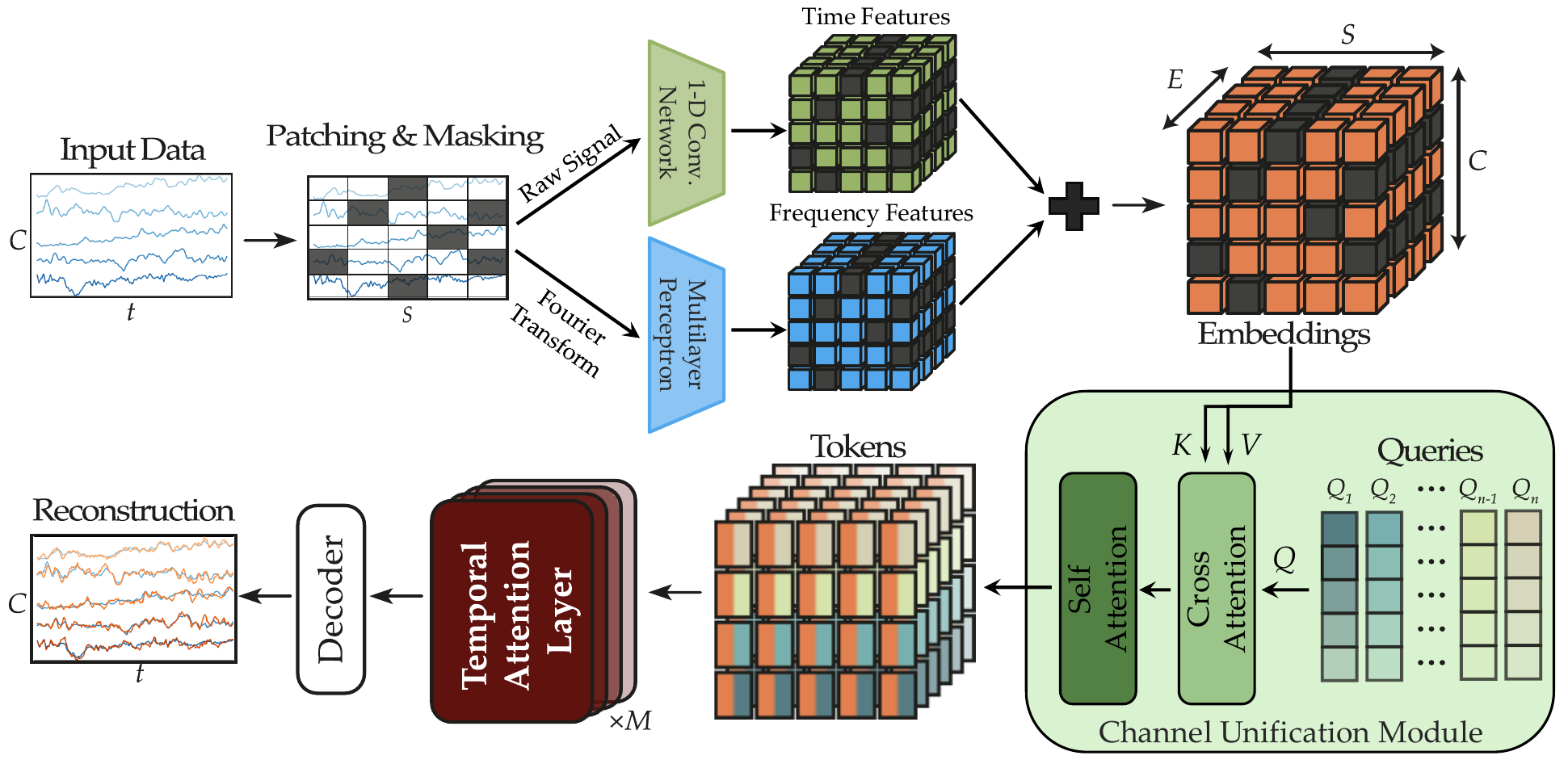}
    \caption{Overview of LUNA. EEG signals ($B \times C \times T$) are segmented into patches and embedded. Channel-Unification Module maps channel-wise features into a fixed-size latent space using learned queries ($Q$). Patch-wise Temporal Attention processes this latent sequence. The decoder generates task-specific outputs.}
    \label{fig:methodlogy}
\end{figure}

LUNA addresses these challenges using a smaller latent space. Firstly, Channel-Unification Module (Sec .~\ref {subsubsec:channel_unification}) employs learned queries and cross-attention to project variable-channel features into a fixed-dimension latent space, achieving topology invariance. Secondly, by unifying channel information into a compact set of \(Q\) queries (\(Q \ll C\)) before temporal processing, LUNA significantly reduces computational demands. This design enables efficient and scalable processing of heterogeneous EEG data, paving the way for more robust foundation models. LUNA adopts an encoder-decoder architecture that transforms EEG signals from heterogeneous montages into a unified latent representation, enabling topology-agnostic modeling and efficient downstream decoding (Figure~\ref{fig:methodlogy}).

\subsection{Encoder}
The encoder comprises three key modules that transform the input EEG into a topology-agnostic latent representation: patch feature extraction, channel unification, and patch-wise temporal modeling.

\paragraph{Patch Feature Extraction}
Given raw EEG \(x \in \mathbb{R}^{B \times C \times T}\) (Batch \(B\), Channels \(C\), Time \(T\)), we segment each channel into \(S = T / P\) non-overlapping temporal patches of size \(P\). These patches are embedded via two parallel pathways:

\textbf{Temporal Embedding:} A 1D convolutional network (with GroupNorm~\citep{wu2018group}, GELU~\citep{hendrycks2016gaussian}) encodes local temporal features similar to state-of-the-art methods such as LaBraM\cite{jiang2024large} and CBraMod \cite{wang2024cbramod},
\textbf{Frequency Embedding:} The magnitude and phase from each patch's Fourier transform are projected through an MLP.
These representations are summed to obtain patch features \(x_\text{features}\).

\vspace{-3mm}
\paragraph{Channel Positional Encoding}
To encode electrode locations, we apply NeRF-inspired sinusoidal encoding~\cite{mildenhallNerf} to normalized 3D electrode coordinates, followed by an MLP projection. This yields $\mathbf{E}_{\text{pos}} \in \mathbb{R}^{B \times C \times E}$, which is added to $x_{\text{features}}$. 

During pre-training, a random subset of tokens is masked using a learnable embedding. 

\paragraph{Channel-Unification Module}
\label{subsubsec:channel_unification}
To handle varying channel counts (\(C\)) across recordings, we introduce a cross-attention module that maps patch-wise features into a fixed latent space. Specifically, Q learned queries $\mathbf{Q}_{\text{learn}} \in \mathbb{R}^{Q \times E}$, (learnable parameters \emph{without} a batch dimension, initialized orthogonally) cross-attend to patch features. For attention, these queries are repeated across the $B \cdot S$ patch instances as $\tilde{\mathbf{Q}} \in \mathbb{R}^{(B \cdot S) \times Q \times E}$.

Let the input to this module be the tensor \( \mathbf{X}_\text{token} \in \mathbb{R}^{B \times (C \cdot S) \times E} \), representing the spatially-aware features for \(B\) samples, \(S\) patches per channel, and feature dimension \(E\). We first reshape this tensor to \( \mathbf{X}' \in \mathbb{R}^{(B \cdot S) \times C \times E} \) to treat each patch instance across the batch independently while isolating the channel dimension for attention.

The cross-attention mechanism then computes the output representation \( \mathbf{A}_{\text{out}} \in \mathbb{R}^{(B \cdot S) \times Q \times E} \):
\begin{equation}
    \mathbf{A}_{\text{out}} = \text{MultiHeadAttention}(\tilde{\mathbf{Q}}, \mathbf{X}', \mathbf{X}')
\end{equation}

A feed-forward network (FFN) with residual connection refines the outputs, followed by $L$ Transformer encoder layers operating on the query dimension $Q$.
\begin{equation}
    \mathbf{X}_{\text{unified}} = \text{TransformerEncoder}(\mathbf{A}_{\text{out}} + \text{FFN}(\mathbf{A}_{\text{out}}))
\end{equation}
The result \( \mathbf{X}_{\text{unified}} \in \mathbb{R}^{(B \cdot S) \times Q \times E} \) decouples further processing from the original electrode layout.  \textit{For clarity: learnable tensors (e.g., $\mathbf{Q}_{\text{learn}}$) omit the batch dimension; repetition occurs only at attention time (e.g., $\tilde{\mathbf{Q}}$).} 

\paragraph{Patch-wise Temporal Encoder}
\label{subsubsec:temporal_encoder}

The unified representations are reshaped into temporal sequences \( \mathbf{X}_{\text{unified}}' \in \mathbb{R}^{B \times S \times (Q \cdot E)} \). These are processed by a stack of Transformer encoder blocks with Rotary Positional Embeddings (RoPE)~\citep{su2024roformer} to capture temporal dependencies efficiently. A key advantage of this encoding approach is that each of the \(S\) temporal tokens in \(\mathbf{X}_{\text{unified}}'\) now encapsulates richer, aggregated information from multiple input channels, rather than representing a single channel's segment. Furthermore, by not tokenizing each channel independently for temporal processing, the effective sequence length for the temporal Transformers is reduced from \(S \cdot C\) to just \(S\), leading to significant reductions in computational complexity and memory requirements. $$E_{\text{out}} = \text{TemporalEncoder}(\mathbf{X}_{\text{unified}}')$$

\subsection{Decoder}
\label{subsec:decoder}
LUNA supports two decoding strategies depending on the task: reconstruction (pre-training) and classification (fine-tuning).

\paragraph{Reconstruction Head (Pre-training)}
For masked patch reconstruction, $C$ learned decoder queries $\mathbf{E}_{\text{dec}}^{\text{learn}} \in \mathbb{R}^{C \times E}$ (learnable parameters \emph{without} a batch dimension) are repeated across the batch and patches as $\tilde{\mathbf{E}}_{\text{dec}} \in \mathbb{R}^{(B \cdot S) \times C \times E}$ and attend to reshaped $E_{\text{out}}$ via cross-attention, producing channel-specific representations $\mathbf{Z}_{\text{dec}} \in \mathbb{R}^{(B\cdot S) \times C \times E}$. $E_{\text{out}} \in \mathbb{R}^{B \times S \times (Q \cdot E)}$ is reshaped to be used as keys/values $\mathbf{K}, \mathbf{V} \in \mathbb{R}^{(B \cdot S) \times Q \times E}$ for attention. A linear projection $\phi:\mathbb{R}^{E}\!\rightarrow\!\mathbb{R}^{P}$ applied on $\mathbf{Z}_{\text{dec}}$  recovers the patch values. 
{\textit{Decoder queries are indexed by channel labels and can be reused across datasets when electrodes overlap; this reconstruction head is used only during pre-training.}

\paragraph{Classification Head (Fine-tuning)}
For downstream tasks, a single aggregation query $\mathbf{E}_{\text{agg}}^{\text{learn}} \in \mathbb{R}^{1 \times (Q \cdot E)}$ (no batch) is repeated across the batch as $\tilde{\mathbf{E}}_{\text{agg}} \in \mathbb{R}^{B \times 1 \times (Q \cdot E)}$ and attends to $E_{\text{out}}$ to produce a pooled representation, which is passed to an MLP for classification.

\subsection{Training Objectives}
\label{subsec:training}
LUNA is pre-trained with a masked reconstruction loss and an auxiliary query specialization loss.

\paragraph{Reconstruction Loss}
A Smooth L1 loss is applied to both masked and visible patches:
$$L_{rec}=\frac{1}{N_\text{masked}} \sum_{i \in M} \text{SmoothL1}(x_{\text{orig}_i}, x_{\text{recons}_i})+\alpha \cdot \frac{1}{N_\text{visible}} \sum_{i \notin M} \text{SmoothL1}(x_{\text{orig}_i}, x_{\text{recons}_i})$$

and SmoothL1($x, \hat{x}$) = $0.5(x - \hat{x})^2$ if $|x - \hat{x}| < \beta$, else $\beta|x - \hat{x}| - 0.5\beta^2$, with $\beta=1$.
\paragraph{Affinity matrix.}
Let $\mathbf{A}_{\text{attn}} \in \mathbb{R}^{B' \times H \times Q \times C}$ denote the cross-attention weights (queries $\rightarrow$ channels) from the channel-unification module, for $B'$ instances and $H$ heads. We define
\[
\mathbf{A}_{\text{affinity}} \;=\; \frac{1}{H}\sum_{h=1}^{H} \mathbf{A}_{\text{attn}}[:,h,:,:] \;\in\; \mathbb{R}^{B' \times Q \times C},
\]
i.e., attention weights averaged over heads, so that $(\mathbf{A}_{\text{affinity}})_{b',q,c}$ measures the affinity between query $q$ and channel $c$ for instance $b'$.
\paragraph{Query Specialization Loss}
To promote diversity among queries, we penalize similarity in query-channel affinity matrices by minimizing the mean value of off-diagonal elements:
\begin{equation*}
    \mathcal{L}_{\text{spec}} = \frac{\lambda_{\text{spec}}}{B' \cdot Q \cdot (Q-1)} \sum_{b'=1}^{B'} \sum_{i=1}^{Q} \sum_{j=1, j \neq i}^{Q} \left( (\mathbf{A}_{\text{affinity}} \mathbf{A}_{\text{affinity}}^T)_{b', i, j} \right)^2
\end{equation*}
\section{Results}
\label{sec:results}
\vspace{-2mm}
\subsection{Experimental Setup}
\label{subsec:setup}
\paragraph{Datasets} 

We pre-train LUNA on a combined corpus of Temple University Hospital EEG Corpus (TUEG)~\citep{obeid2016temple} and the Siena Scalp EEG Database \citep{detti2020siena}, spanning recordings with 20, 22, and 29 channels amounting to over 21,900 hours of EEG data (see Table~\ref{tab:app_dataset_summary}). Downstream evaluations cover four diverse benchmarks: \textbf{TUAB} \citep{obeid2016temple}: Abnormal EEG detection (binary classification), \textbf{TUAR} \citep{obeid2016temple}: We follow prior work~\citep{ingolfsson2022} and treat artifact detection as a \emph{multiclass} (single-label) problem with five classes (one label per segment). Artifact detection (multi-class classification) \textbf{TUSL} \citep{obeid2016temple}: Slowing event classification (4-class classification). \textbf{SEED-V} \citep{liu2021comparing}: Emotion recognition (5-class classification), with unseen 62-channel topology. All subjects and recordings from the downstream evaluation datasets (TUAB, TUAR, TUSL, SEED-V) were strictly excluded from this pre-training set to ensure fair evaluation of generalization. For LUNA, the input EEG is segmented into patches, consisting of 40 timestamps. For most datasets, EEG recordings are sliced into non-overlapping 5-second segments to form individual training/evaluation samples. SEED-V dataset uses its default 1-second sample duration.

\paragraph{Fine-tuning and Data Splits}
For the TUAB dataset, we use the official train-test split. As TUSL and TUAR lack official subject-wise test splits, we follow recent leading work (e.g., EEGFormer~\citep{chen2024eegformer}) and adopt an 80\%/10\%/10\% randomized sample-level split for train/val/test to allow direct, like-for-like comparison. We acknowledge that subject-independent splits are the gold standard for assessing clinical generalization and recommend them for future benchmark comparisons. For SEED-V, fifteen trials are divided equally into train, validation, and test sets for each session. For the TUAR dataset, we adopt a multiclass classification approach, restricting to 5 distinct artifact types in a single-label setting, similar to EEGFormer~\citep{chen2024eegformer}. We optimize binary cross-entropy loss for TUAB and cross-entropy loss for other datasets. We report the mean and standard deviation of results obtained across three different random seeds.

\paragraph{Preprocessing}
We apply a minimal, standardized preprocessing pipeline to all EEG data. Signals are first bandpass filtered between 0.1 Hz and 75 Hz. A notch filter (50Hz or 60Hz) is applied to remove power-line interference. All signals are then resampled to 256 Hz. 
For TUEG, TUAB, TUAR, and TUSL datasets we construct a bipolar (“double-banana”) montage by differencing predefined longitudinal electrode pairs provided in the dataset documentation; the full list of channel pairs used is given in Appendix~\ref{app:bipolar_pairs}. Siena and SEED-V are processed in unipolar format. Finally, each channel within each sample is normalized using z-score normalization.

\paragraph{Computational Environment}
All experiments were conducted on a cluster of eight NVIDIA A100 GPUs, using Python 3.11.6 and PyTorch 2.4.1 with CUDA 12.1. Training utilizes `bf16` mixed-precision. Detailed hyperparameters for pre-training and fine-tuning are provided in Appendix~\ref{app:exp_settings}.

\paragraph{Baselines and Variants} 
We compare against state-of-the-art supervised and self-supervised methods, including transformer-based architectures such as LaBraM~\citep{jiang2024large}, CBraMod~\citep{wang2024cbramod}, EEGFormer~\citep{chen2024eegformer}, and BIOT~\citep{yang2023biot}. LUNA is evaluated in three configurations: Base (7M), Large (43M), and Huge (311M parameters). Model size is increased by expanding the depth of the Patch-wise Temporal Encoder, the hidden embedding dimension \(E\), and the number/size of queries \(Q\) in the Channel-Unification Module. Key architectural settings are detailed in Appendix~\ref{app:model_details}.

\subsection{Downstream Task Performance}
\label{subsec:downstream_results}

\paragraph{Abnormal EEG Detection (TUAB)}
LUNA delivers competitive performance on TUAB (Table \ref{tab:results_tuab}). LUNA-Huge achieves AUROC of 0.8957 and AUPR of 0.9029, surpassing most self-supervised baselines and approaching large-scale models like LaBraM and CBraMod. Notably, LUNA maintains strong performance while offering substantial efficiency and topology-agnostic benefits relative to strong self-supervised and large-scale baselines.

\begin{table}[htbp!]
    \centering
    \caption{Performance comparison on TUAB abnormal EEG detection.}
    \label{tab:results_tuab}
    
    \setlength{\tabcolsep}{6pt} 
    \small
    \begin{tabular}{@{}lcccc@{}}
        \toprule
        \textbf{Model} & \textbf{Size} & \textbf{Bal. Acc. (\%)} $\uparrow$ & \textbf{AUC-PR} $\uparrow$ & \textbf{AUROC} $\uparrow$ \\
        \midrule
        \multicolumn{5}{l}{\textit{Supervised Models}} \\    SPaRCNet~\cite{jing2023development} & 0.8M & 78.96 $\pm$ 0.18 & 0.8414 $\pm$ 0.0018 & 0.8676 $\pm$ 0.0012 \\
        ContraWR~\cite{yang2021self} & 1.6M & 77.46 $\pm$ 0.41 & 0.8421 $\pm$ 0.0140 & 0.8456 $\pm$ 0.0074 \\
        CNN-Transformer~\cite{peh2022transformer} & 3.2M & 77.77 $\pm$ 0.22 & 0.8433 $\pm$ 0.0039 & 0.8461 $\pm$ 0.0013 \\
        FFCL~\cite{li2022motor} & 2.4M & 78.48 $\pm$ 0.38 & 0.8448 $\pm$ 0.0065 & 0.8569 $\pm$ 0.0051 \\
        ST-Transformer~\cite{song2021transformer} & 3.2M & 79.66 $\pm$ 0.23 & 0.8521 $\pm$ 0.0026 & 0.8707 $\pm$ 0.0019 \\
        \midrule
        \multicolumn{5}{l}{\textit{Self-supervised Models}} \\
        BENDR \citep{kostas2021bendr} & 0.39M & 76.96 $\pm$ 3.98 & -  & 0.8397 $\pm$ 0.0344 \\
        BrainBERT \citep{wang2023brainbert} & 43.2M & - & 0.8460 $\pm$ 0.0030 & 0.8530 $\pm$ 0.0020 \\
        EEGFormer-Base \citep{chen2024eegformer} & 2.3M & - & 0.8670 $\pm$ 0.0020 & 0.8670 $\pm$ 0.0030 \\
        BIOT \citep{yang2023biot} & 3.2M & 79.59 $\pm$ 0.57 & 0.8692 $\pm$ 0.0023 & 0.8815 $\pm$ 0.0043 \\
    EEG2Rep~\cite{mohammadi2024eeg2rep} & - & 80.52 $\pm$ 2.22 & - & 0.8843 $\pm$ 0.0309 \\
        FEMBA-Huge~\cite{tegon2025femba} & 386M  & 81.82 $\pm$ 0.16 & 0.9005 $\pm$ 0.0017 & 0.8921 $\pm$ 0.0042  \\
        CEReBrO~\cite{dimofte2025cerebro} & 85.15M & 81.67 $\pm$ 0.23 & 0.9049 $\pm$ 0.0026 & 0.8916 $\pm$ 0.0038 \\
        LaBraM-Base \citep{jiang2024large} & 5.9M & 81.40 $\pm$ 0.19 & 0.8965 $\pm$ 0.0016 & 0.9022 $\pm$ 0.0009 \\
        LaBraM-Huge \citep{jiang2024large} & 369.8M & \textbf{82.58 $\pm$ 0.11} & 0.9204 $\pm$ 0.0011 & \textbf{0.9162 $\pm$ 0.0016} \\
        CBraMod \citep{wang2024cbramod} & 69.3M & 82.49 $\pm$ 0.25 & \textbf{0.9221 $\pm$ 0.0015} & 0.9156 $\pm$ 0.0017 \\
        \midrule
        \textbf{LUNA-Base} & 7M & 80.63 $\pm$ 0.08 & 0.8953 $\pm$ 0.0016 & 0.8868 $\pm$ 0.0015 \\
        \textbf{LUNA-Large} & 43M & 80.96 $\pm$ 0.10 & 0.8986 $\pm$ 0.0005 & 0.8924 $\pm$ 0.0010 \\
        \textbf{LUNA-Huge} & 311.4M & 81.57 $\pm$ 0.11 & 0.9029 $\pm$ 0.0014 & 0.8957 $\pm$ 0.0011 \\
        \bottomrule
   \end{tabular}
\end{table}
\paragraph{Artifact and Slowing Detection (TUAR and TUSL)}
LUNA delivers state-of-the-art results on TUAR and TUSL (Table \ref{tab:results_tuar_tusl}). LUNA-Huge achieves AUROC 0.921 on TUAR, outperforming FEMBA-Large and other methods. On TUSL, LUNA-Huge reaches AUROC 0.802, the highest among all compared models. 

\begin{table}[htbp!]
\centering
\caption{Performance comparison on TUAR (artifact detection) and TUSL (slowing event classification). }
\label{tab:results_tuar_tusl}

\setlength{\tabcolsep}{5pt}
\small
\begin{tabular}{@{}lccccccc@{}}
\toprule
\multirow{2}{*}{\textbf{Model}} & \multirow{2}{*}{\textbf{Size}} & \multicolumn{2}{c}{\textbf{TUAR}} & \multicolumn{2}{c}{\textbf{TUSL}} \\
\cmidrule(lr){3-4} \cmidrule(lr){5-6}
 &  & \textbf{AUROC} $\uparrow$ & \textbf{AUC-PR} $\uparrow$ & \textbf{AUROC} $\uparrow$ & \textbf{AUC-PR} $\uparrow$ \\
\midrule
\multicolumn{6}{l}{\textit{Supervised Models}} \\
EEGNet~\cite{lawhern2018eegnet}       & -  & 0.752 $\pm$ 0.006 & 0.433 $\pm$ 0.025 & 0.635 $\pm$ 0.015 & 0.351 $\pm$ 0.006  \\
EEG-GNN~\cite{tang2021self}      & - & 0.837 $\pm$ 0.022 & 0.488 $\pm$ 0.015 & 0.721 $\pm$ 0.009 & 0.381 $\pm$ 0.004  \\
GraphS4mer~\cite{tang2023modeling}  & -  & 0.833 $\pm$ 0.006 & 0.461 $\pm$ 0.024 & 0.632 $\pm$ 0.017 & 0.359 $\pm$ 0.001  \\
\midrule
\multicolumn{6}{l}{\textit{Self-supervised Models}} \\
BrainBERT \citep{wang2023brainbert}    & 43.2M  & 0.753 $\pm$ 0.012 & 0.350 $\pm$ 0.014 & 0.588 $\pm$ 0.013 & 0.352 $\pm$ 0.003  \\
EEGFormer-Base \citep{chen2024eegformer}  & 2.3M & 0.847 $\pm$ 0.014 & 0.483 $\pm$ 0.026 & 0.713 $\pm$ 0.010 & \textbf{0.393 $\pm$ 0.003}  \\
EEGFormer-Large \citep{chen2024eegformer}  & 3.2M & 0.852 $\pm$ 0.004 & 0.483 $\pm$ 0.014 & 0.679 $\pm$ 0.013 & 0.389 $\pm$ 0.003  \\
FEMBA-Base \citep{tegon2025femba}  & 47.7M & 0.900 $\pm$ 0.010 & \textbf{0.559 $\pm$ 0.002} & 0.731 $\pm$ 0.012 & 0.289 $\pm$ 0.009 \\
FEMBA-Large \citep{tegon2025femba} & 77.8M  & 0.915 $\pm$ 0.003 & 0.521 $\pm$ 0.001 & 0.714 $\pm$ 0.007 & 0.282 $\pm$ 0.010 \\
\midrule
\textbf{LUNA-Base} & 7M & 0.902 $\pm$ 0.011 	 & 0.495 $\pm$	0.010 & 0.767 $\pm$	0.023 & 0.301 $\pm$	0.003 \\
\textbf{LUNA-Large} & 43M & 0.918 $\pm$	0.003	& 0.505 $\pm$	0.010 & 0.771	$\pm$ 0.006	& 0.293	$\pm $0.021  \\
\textbf{LUNA-Huge} & 311.4M & \textbf{0.921 $\pm$	0.011}	 & 0.528 $\pm$	0.012 & \textbf{0.802 $\pm$	0.005}	& 0.289	$\pm $0.008  \\
\bottomrule
\end{tabular}

\end{table}

\paragraph{Emotion Recognition on Unseen Montage (SEED-V)}
The SEED-V benchmark tests generalization to a novel 62-channel montage, distinct from pre-training data. Results in Table \ref{tab:emotion_classification} show that while LUNA effectively operates on this unseen topology, its performance (e.g., Bal. Acc.) lags behind leading methods like CBraMod by 2-3 pp. This suggests a trade-off inherent in LUNA's design: while its query-based unification enables efficient, topology-agnostic processing across common montage variations (as demonstrated on TUAB/TUAR/TUSL), generalizing zero-shot to vastly different, high-density layouts remains challenging, possibly due to positional encoding constraints. Despite this gap, LUNA shows positive scaling from Base to Large models, underscoring its potential.

\vspace{-3mm}
\begin{table}[htbp!]
    \centering
    \caption{Performance comparison on SEED-V emotion recognition (5-classes).}
    \label{tab:emotion_classification}
    \small
    \begin{tabular}{@{}lcccc@{}}
        \toprule
        \textbf{Model} & \textbf{Size} & \textbf{Bal. Acc. (\%)} $\uparrow$ & \textbf{Cohen’s Kappa } $\uparrow$ & \textbf{Weighted F1} $\uparrow$ \\
        \midrule
         \multicolumn{5}{l}{\textit{Supervised Models}} \\
        SPaRCNet~\cite{jing2023development} &0.79M&0.2949 $\pm$ 0.0078 &0.1121 $\pm$ 0.0139 &0.2979 $\pm$ 0.0083 \\
        ContraWR~\cite{yang2021self}&1.6M&0.3546 $\pm$ 0.0105 &0.1905 $\pm$ 0.0188 &0.3544 $\pm$ 0.0121 \\
        CNN-Transformer~\cite{peh2022transformer}&3.2M&0.3678 $\pm$ 0.0078& 0.2072 $\pm$ 0.0183&0.3642 $\pm$ 0.0088 \\
        FFCL~\cite{li2022motor}  &    2.4M &0.3641 $\pm$ 0.0092 &0.2078 $\pm$ 0.0201&0.3645 $\pm$ 0.0132\\
        ST-Transformer~\cite{song2021transformer} &3.5M &0.3052 $\pm$ 0.0072 &0.1083 $\pm$ 0.0121 &0.2833 $\pm$ 0.0105 \\
						\midrule
        \multicolumn{5}{l}{\textit{Self-supervised Models}} \\
        BIOT \citep{yang2023biot} & 3.2M & 0.3837 $\pm$ 0.0187 & 0.2261 $\pm$ 0.0262 &  0.3856 $\pm$ 0.0203 \\
        LaBraM-Base \citep{jiang2024large} & 5.8M & 0.3976 $\pm$ 0.0138 & 0.2386 $\pm$ 0.0209 & 0.3974 $\pm$ 0.0111 \\
        CBraMod \citep{wang2024cbramod} & 14M & \textbf{0.4091 $\pm$ 0.0097} & \textbf{0.2569 $\pm$ 0.0151} & \textbf{0.4101 $\pm$ 0.0108} \\
        \midrule
        
        \textbf{LUNA-Base} & 7M & 0.3730 $\pm$ 0.0098 & 0.1831 $\pm$ 0.0103 & 0.3389 $\pm$ 0.0091 \\
        \textbf{LUNA-Large} & 43M & 0.3918 $\pm$ 0.0066 & 0.2073 $\pm$ 0.0045 & 0.3586 $\pm$ 0.0013 \\
        \textbf{LUNA-Huge} & 311.4M & 0.3900 $\pm$ 0.0096 & 0.2037 $\pm$ 0.0103 & 0.3506 $\pm$ 0.0047 \\
        \bottomrule
    \end{tabular}
\end{table}
Unless noted, we report mean~$\pm$~s.d.\ over matched seeds and focus on effect sizes and confidence intervals. Formal significance tests are summarized in Appendix~\ref{app:sig_tests}

\subsection{Computational Efficiency}
\label{subsec:efficiency}

\textbf{LUNA achieves substantially better calling efficiency} compared to full and alternating attention models. As shown in Figure~\ref{fig:patch_scaling}, LUNA's patch-wise attention enables thousands of temporal patches without the quadratic cost faced by LaBraM. Likewise, Figure~\ref{fig:channel_scaling} shows that LUNA maintains near-constant compute cost when channel count increases, outperforming CBraMod's $\bigO(C^2)$ scaling for dense EEG recordings. These results confirm that LUNA decouples inference cost from input montage, making it well-suited for long recordings or high-density EEG scenarios. We also consider BIOT (linear attention) as an efficiency-oriented baseline; detailed FLOPs/memory scaling versus LUNA is provided in Appendix~\ref{app:biot_scaling}. Although many public datasets use $\sim$20–30 channels, research/clinical systems often employ 64–256 channels and longer windows; LUNA’s $O(C)$ unification and reduced temporal sequence length enable such regimes where quadratic spatial attention becomes impractical.

\begin{figure}[htbp!]
    \centering
    \begin{subfigure}{\linewidth} 
        \centering
        \includegraphics[width=\linewidth]{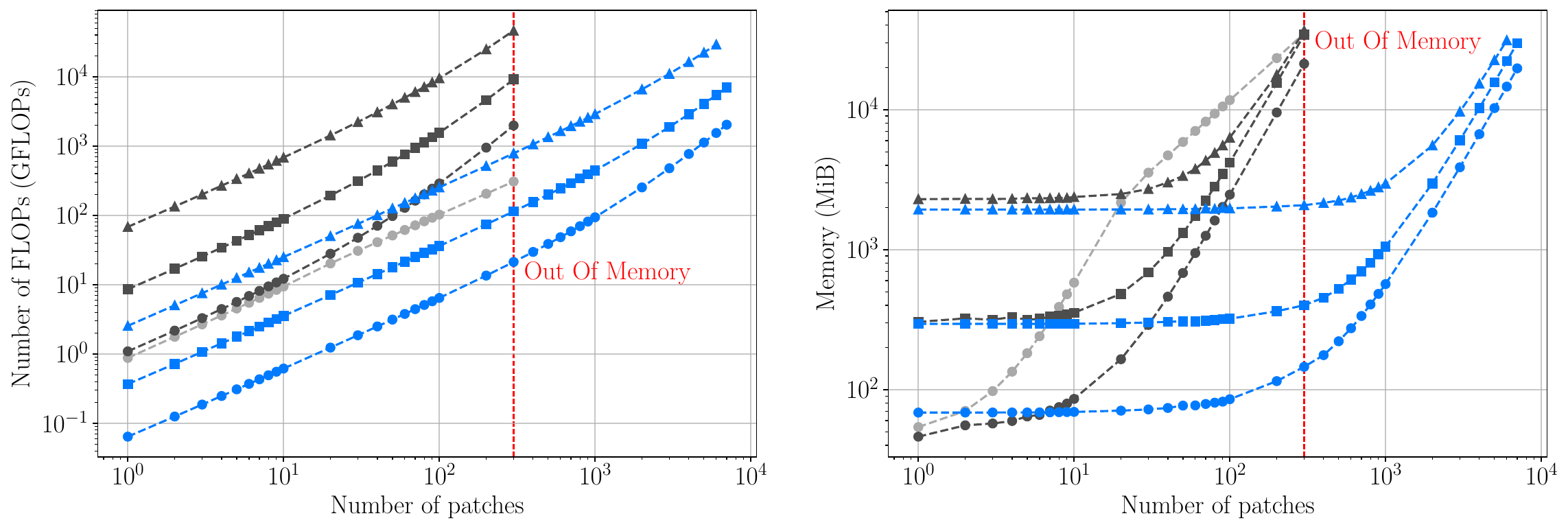}
        \caption{Scaling with number of patches (Channels fixed at 22).}
        \label{fig:patch_scaling}
    \end{subfigure}
    \vspace{-0.2cm}

    \begin{subfigure}{\linewidth}
        \centering
        \includegraphics[width=\linewidth]{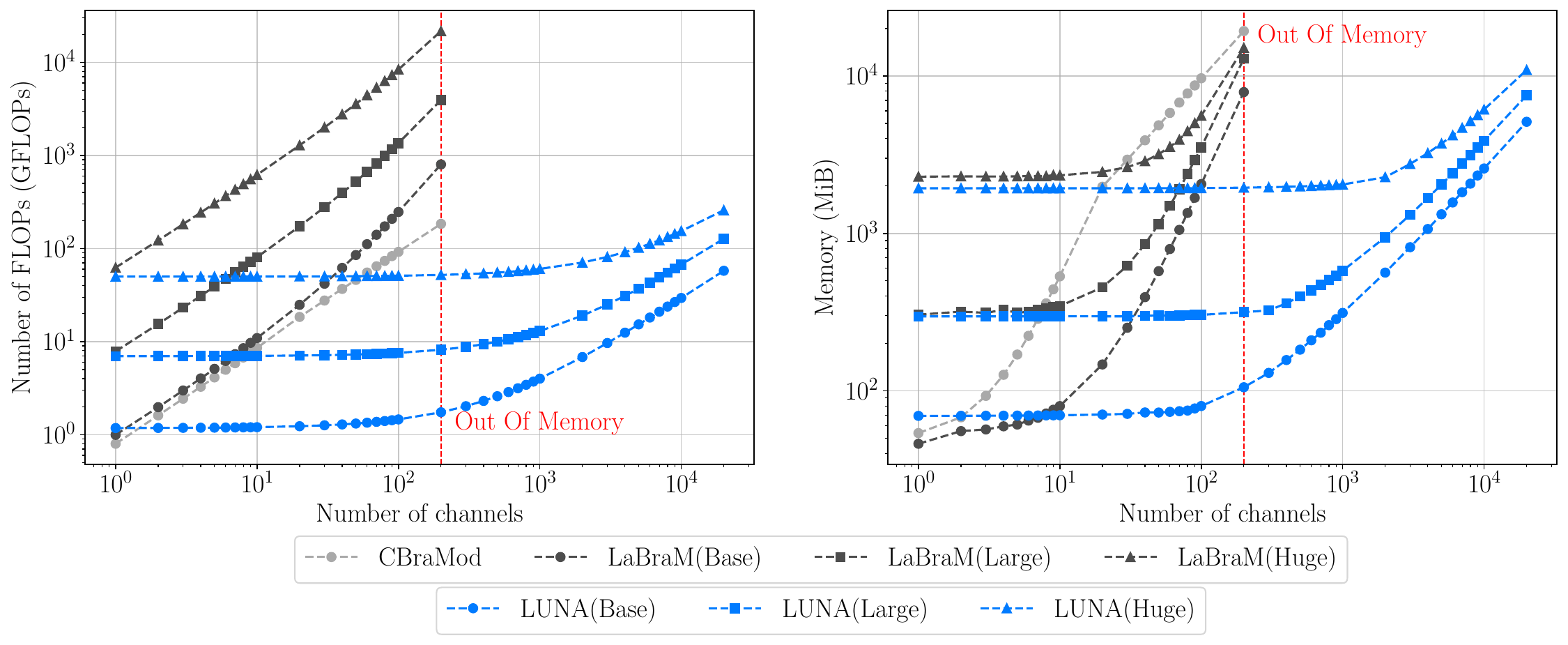}
        \caption{Scaling with number of channels (Patches fixed at 20).}
    \label{fig:channel_scaling}
    \end{subfigure}
    \vspace{-0.6cm}
    \caption{Computational cost scaling of LUNA and baseline models. (a) FLOPs and Memory usage vs. number of patches. (b) FLOPs and Memory usage vs. number of channels. LUNA demonstrates significantly better efficiency and scalability, especially compared to full attention (LaBraM), and favorable scaling compared to alternating attention (CBraMod) due to the fixed latent query space. CBraMod has a variable sized decoder based on the number of patches and channels; therefore, its model size as well as its resource usage grows rapidly. \emph{FLOPs are measured with \texttt{fvcore}'s \texttt{FlopCountAnalysis} over 50 random inputs, including encoder{+}decoder, with window length $T$, patch size $P$, and reported as GFLOPs per forward pass.}}
    \vspace{-0.7cm}
\label{fig:scaling_plots_combined}
\end{figure}

\subsection{Ablation Studies}
\label{subsec:ablations}
\emph{Choice of $Q$.} We explore the trade-off between the number of queries $Q$ and embedding size $E$ under a fixed $Q\!\cdot\!E$ budget; see Appendix~\ref{app:q_tradeoff}. We validate the impact of LUNA's key design choices on TUAB and TUAR (Table \ref{tab:ablation}).

\vspace{-4mm}
\paragraph{Learned Queries vs. Fixed Regions}
Replacing learned queries with predefined spatial regions (similar to MMM~\citep{yi2023learning}) yields small AUROC changes ($-0.004$ to $-0.006$), within seed variation. We therefore emphasize the \emph{practical} advantages of learned queries—data-driven flexibility without anatomy-specific priors—rather than a statistically significant metric gain (Appendix~\ref{app:sig_tests}).

\vspace{-4mm}
\paragraph{Query Specialization Loss}
Removing the specialization loss results in modest AUROC changes (-0.003 to -0.006), again on the order of the reported variation, with small mixed effects on AUC-PR. We retain this loss for its \emph{regularizing role}: it encourages a diverse, non-redundant set of spatial filters (see Fig.~\ref{fig:query_analysis}), which is desirable for robustness in challenging artifact conditions.

\vspace{-4mm}
\paragraph{Frequency Features}
Ablating frequency embeddings leads to the largest drop (up to -0.012 AUROC), indicating a more consistent contribution complementary to temporal features.

\subsection{Latent Space Analysis}
\paragraph{Pre-trained Representations}
t-SNE visualizations (Figure~\ref{fig:app_tsne_plots}) reveal that even before fine-tuning, LUNA's encoder captures task-relevant structure. Normal and abnormal EEGs form separate clusters in TUAB, while artifact classes are partially separated in TUAR, demonstrating effective pre-training.

\begin{figure}[htbp!]
    \centering
    \begin{subfigure}[b]{0.48\textwidth}
        \centering
        \includegraphics[width=\linewidth]{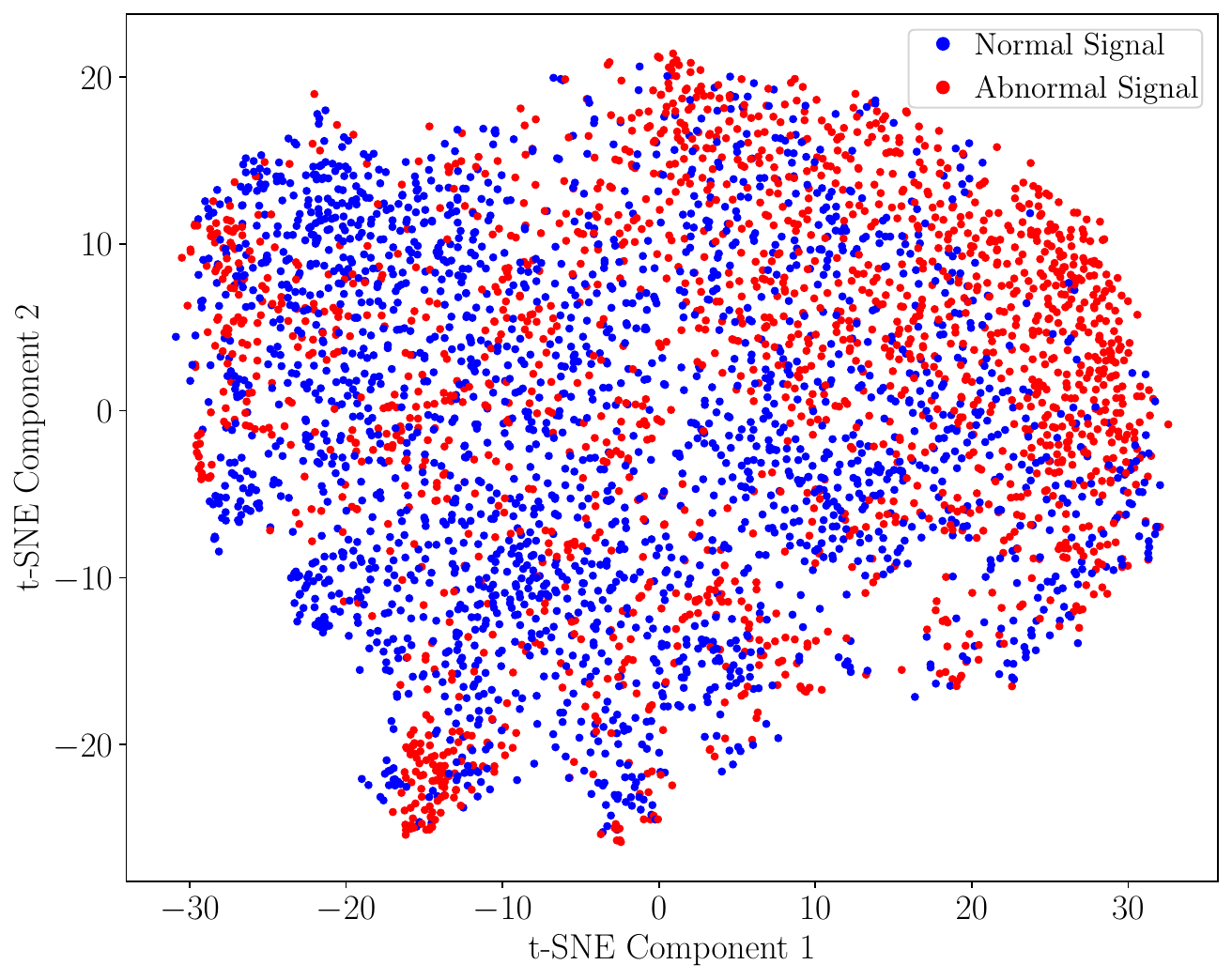}
        \caption{TUAB dataset (Normal vs. Abnormal Signal).}
        \label{fig:app_TUAB_emb}
    \end{subfigure}
    \hfill 
    \begin{subfigure}[b]{0.48\textwidth}
        \centering
        \includegraphics[width=\linewidth]{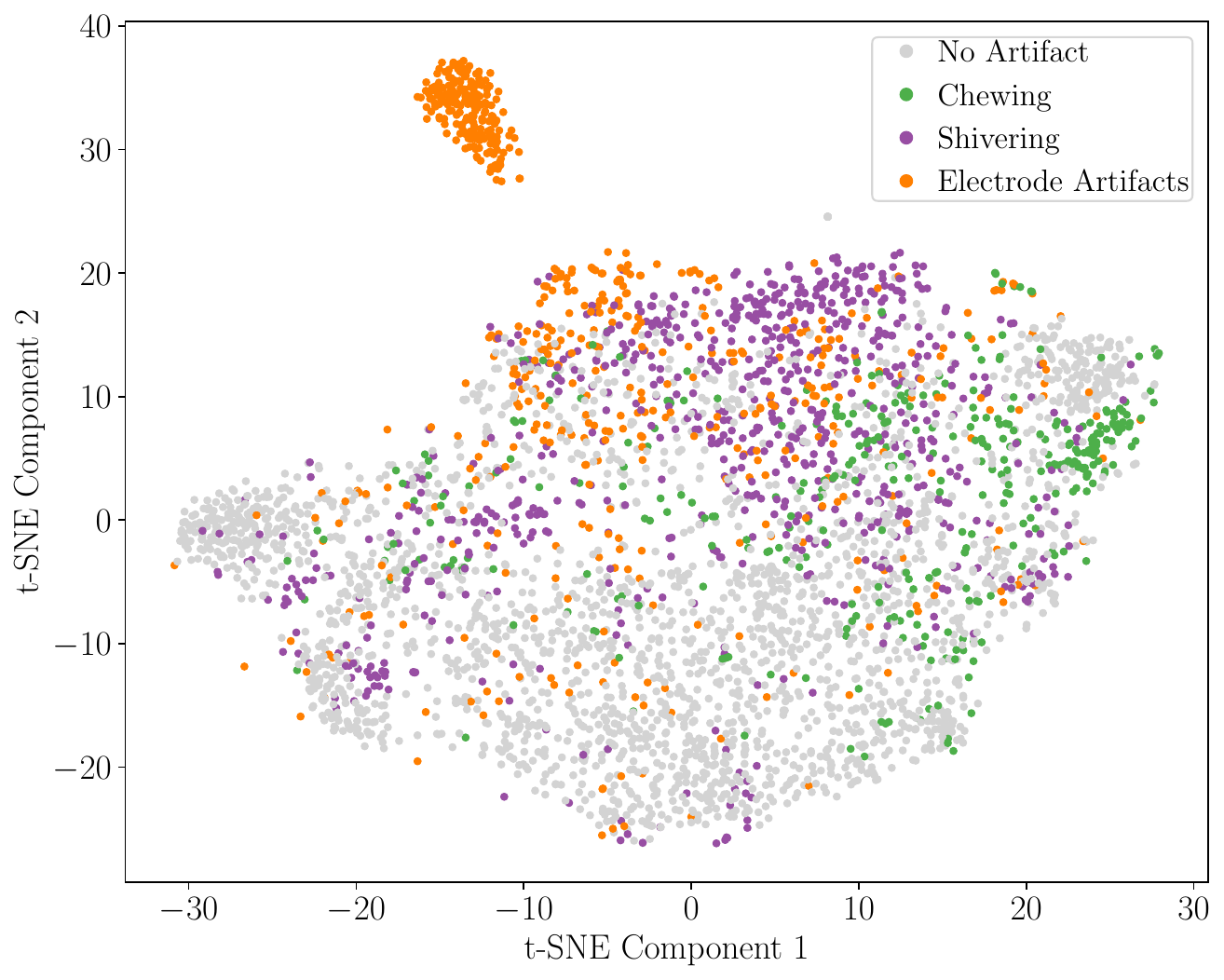} 
        \caption{TUAR dataset (Artifact Types).}
        \label{fig:app_TUAR_emb}
    \end{subfigure}
    \caption{t-SNE of LUNA-Base embeddings on downstream datasets before fine-tuning.}
    \label{fig:app_tsne_plots} 
    \vspace{-0.5cm}
\end{figure}

\subsection{Learned Query Specialization Visualization}
\label{subsec:qualitative}

\paragraph{Query Specialization}
Visual analysis of the learned queries (Figure~\ref{fig:query_analysis}) highlights their role in topology-agnostic representation. Queries exhibit distinct spatial profiles: some are localized (e.g., frontal regions), while others aggregate broader signals. This emergent specialization confirms that cross-attention learns flexible, data-driven basis functions for spatial unification.

\begin{figure}[htbp!]
    \vspace{-0.2cm}
    \centering
    \includegraphics[width=1\linewidth]{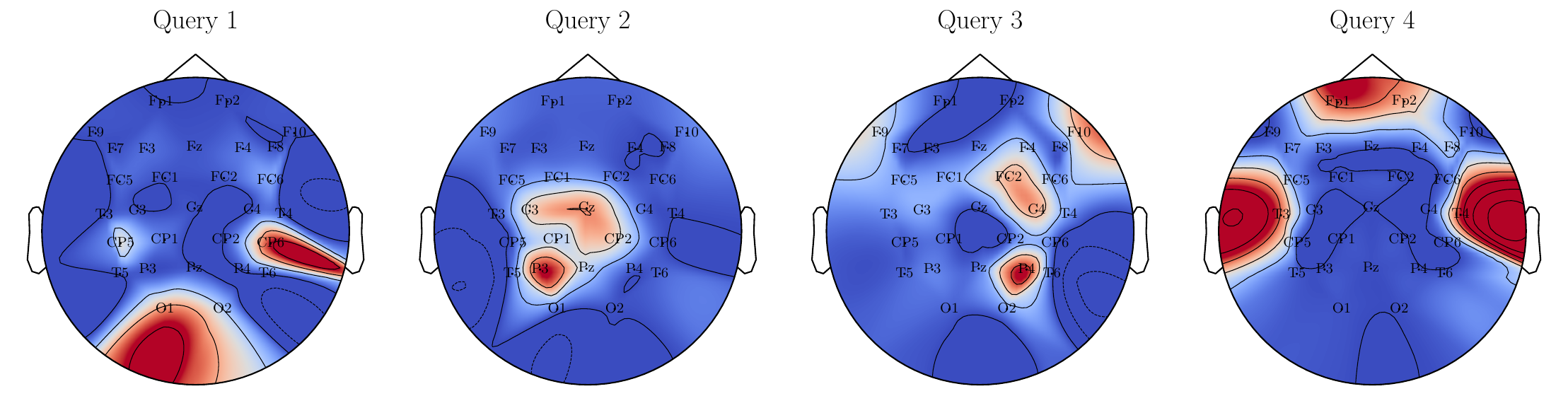}
    \caption{Visualization of the attention patterns of queries in LUNA-Base on Siena~\citep{detti2020siena} topology.}
    \label{fig:query_analysis}
    \vspace{-0.2cm}
\end{figure}

\begin{table}[htbp!]
    \centering
    \vspace{-0.4cm}
   \caption{Ablation study results (LUNA-Base) on TUAB and TUAR datasets.}
    \label{tab:ablation}
    \scriptsize
    \setlength{\tabcolsep}{2pt}
    \begin{tabular}{@{}lcccc@{}}
        \toprule
        \textbf{Model Configuration} & \textbf{TUAB AUROC} & \textbf{TUAB AUC-PR} & \textbf{TUAR AUROC} & \textbf{TUAR AUC-PR} \\
        \midrule  
        LUNA-Base (Full Model) & 0.887 $\pm$ 0.002 & 0.895 $\pm$ 0.002 & 0.902 $\pm$ 0.011 & 0.495 $\pm$ 0.010 \\
        \midrule
        \textit{Unification Module:} \\
        \quad - Region-based Attention  & 0.883 $\pm$ 0.001 \textcolor{red}{($\downarrow$ 0.004)} & 0.892 $\pm$ 0.002 \textcolor{red}{($\downarrow$ 0.003)} & 0.896 $\pm$ 0.001 \textcolor{red}{($\downarrow$ 0.006)} & 0.509 $\pm$ 0.006 \textcolor{green}{($\uparrow$ 0.014)} \\
        \midrule
        \textit{Other Components:} \\
        \quad - w/o Query Specialization Loss & 0.884 $\pm$ 0.003 \textcolor{red}{($\downarrow$ 0.003)} & 0.892 $\pm$ 0.002 \textcolor{red}{($\downarrow$ 0.003)} & 0.895 $\pm$ 0.005 \textcolor{red}{($\downarrow$ 0.007)} & 0.498 $\pm$ 0.010 \textcolor{green}{($\uparrow$ 0.003)} \\

        \quad - w/o Frequency Features & 0.876 $\pm$ 0.012 \textcolor{red}{($\downarrow$ 0.011)} & 0.883 $\pm$ 0.005 \textcolor{red}{($\downarrow$ 0.012)} & 0.893 $\pm$ 0.011 \textcolor{red}{($\downarrow$ 0.009)} & 0.490 $\pm$ 0.011 \textcolor{red}{($\downarrow$ 0.005)} \\
        \bottomrule
    \end{tabular}
    \vspace{-0.4cm}
\end{table}
\section{Conclusion}
\label{sec:conclusion}

We introduced \textbf{LUNA}, a self-supervised foundation model designed to address the challenge of topological heterogeneity in EEG analysis. By leveraging learned queries and cross-attention, LUNA unifies recordings with diverse electrode layouts into a fixed latent space, enabling montage-agnostic modeling. Through extensive experiments across abnormality detection, artifact recognition, slowing classification, and emotion recognition, we demonstrate that LUNA matches or surpasses state-of-the-art performance while offering substantial efficiency gains in FLOPs and memory usage. Critically, these benefits hold across all evaluated electrode configurations.

While LUNA achieves strong results, especially on heterogeneous montages, our analysis also reveals limitations. Performance on SEED-V suggests sensitivity to unseen channel topologies, likely stemming from reliance on positional encodings learned during pre-training. Addressing this limitation, through enhanced spatial generalization strategies or hybrid learned/geometric embeddings, is an important direction for future work.

More broadly, this work highlights the promise of topology-agnostic latent representations for scalable EEG modeling. Future extensions include exploring unified models across EEG and invasive modalities (e.g., sEEG, ECoG), integrating domain-specific priors (e.g., neurophysiological constraints), and adapting LUNA for real-time inference scenarios. Beyond technical advancements, the development of efficient, topology-invariant EEG models like LUNA could enhance neurological diagnostics and research accessibility. However, careful attention must be paid to mitigating risks such as algorithmic bias and ensuring patient data privacy for deployment. Future work should integrate ethical concerns alongside technical improvements. Pre-training montage diversity is limited (three dominant layouts); future work will incorporate multi-dataset pre-training and randomized channel dropout to improve generalization to unseen, dense and sparse montages.

\begin{ack}
This project is supported by the Swiss National Science Foundation under the grant number 193813 (PEDESITE project). This work was supported by the ETH Future Computing Laboratory (EFCL), financed by a donation from Huawei Technologies. This work was supported by a grant from the Swiss National Supercomputing Centre (CSCS) under project ID lp12 on Alps.
\end{ack}
\bibliographystyle{unsrt}
\bibliography{updated_ref}

\begin{thebibliography}{10}

\bibitem{craik2019deep}
Alexander Craik, Yongtian He, and Jose~L Contreras-Vidal.
\newblock Deep learning for electroencephalogram ({EEG}) classification tasks: a review.
\newblock {\em Journal of Neural Engineering}, 16(3):031001, April 2019.

\bibitem{song2022eeg}
Yonghao Song, Qingqing Zheng, Bingchuan Liu, and Xiaorong Gao.
\newblock {EEG} conformer: Convolutional transformer for {EEG} decoding and visualization.
\newblock {\em IEEE Transactions on Neural Systems and Rehabilitation Engineering}, 31:710--719, December 2023.

\bibitem{wen2022transformers}
Qingsong Wen, Tian Zhou, Chaoli Zhang, Weiqi Chen, Ziqing Ma, Junchi Yan, and Liang Sun.
\newblock Transformers in time series: A survey.
\newblock In {\em Proceedings of the Thirty-Second International Joint Conference on Artificial Intelligence}, IJCAI-2023, pages 6778--6786. International Joint Conferences on Artificial Intelligence Organization, August 2023.

\bibitem{xu2020cross}
Lichao Xu, Minpeng Xu, Yufeng Ke, Xingwei An, Shuang Liu, and Dong Ming.
\newblock Cross-dataset variability problem in {EEG} decoding with deep learning.
\newblock {\em Frontiers in Human Neuroscience}, 14:103, April 2020.

\bibitem{yang2023biot}
Chaoqi Yang, M~Westover, and Jimeng Sun.
\newblock {BIOT}: Biosignal transformer for cross-data learning in the wild.
\newblock In A.~Oh, T.~Naumann, A.~Globerson, K.~Saenko, M.~Hardt, and S.~Levine, editors, {\em Advances in Neural Information Processing Systems}, volume~36, pages 78240--78260. Curran Associates, Inc., September 2023.

\bibitem{yi2023learning}
Ke~Yi, Yansen Wang, Kan Ren, and Dongsheng Li.
\newblock Learning topology-agnostic {EEG} representations with geometry-aware modeling.
\newblock In {\em Thirty-seventh Conference on Neural Information Processing Systems}, September 2023.

\bibitem{mentzelopoulos2024neural}
Georgios Mentzelopoulos, Evangelos Chatzipantazis, Ashwin~G Ramayya, Michelle Hedlund, Vivek Buch, Kostas Daniilidis, Konrad Kording, and Flavia Vitale.
\newblock Neural decoding from stereotactic {EEG}: accounting for electrode variability across subjects.
\newblock In {\em The Thirty-eighth Annual Conference on Neural Information Processing Systems}, September 2024.

\bibitem{lin2023eeg}
Xuefen Lin, Jielin Chen, Weifeng Ma, Wei Tang, and Yuchen Wang.
\newblock {EEG} emotion recognition using improved graph neural network with channel selection.
\newblock {\em Computer Methods and Programs in Biomedicine}, 231:107380, April 2023.

\bibitem{obeid2016temple}
Iyad Obeid and Joseph Picone.
\newblock The temple university hospital {EEG} data corpus.
\newblock {\em Frontiers in Neuroscience}, 10:196, May 2016.

\bibitem{detti2020siena}
Paolo Detti, Giampaolo Vatti, and Garazi Zabalo Manrique~de Lara.
\newblock Siena scalp {EEG} database, 2020.
\newblock This work is licensed under the Creative Commons Attribution 4.0 International License. To view a copy of this license, visit \url{http://creativecommons.org/licenses/by/4.0/}.

\bibitem{liu2021comparing}
Wei Liu, Jie-Lin Qiu, Wei-Long Zheng, and Bao-Liang Lu.
\newblock Comparing recognition performance and robustness of multimodal deep learning models for multimodal emotion recognition.
\newblock {\em IEEE Transactions on Cognitive and Developmental Systems}, 14(2):715--729, June 2022.

\bibitem{kostas2021bendr}
Demetres Kostas, Stephane Aroca-Ouellette, and Frank Rudzicz.
\newblock {BENDR}: Using transformers and a contrastive self-supervised learning task to learn from massive amounts of {EEG} data.
\newblock {\em Frontiers in Human Neuroscience}, 15:653659, June 2021.

\bibitem{wang2023brainbert}
Christopher Wang, Vighnesh Subramaniam, Adam~Uri Yaari, Gabriel Kreiman, Boris Katz, Ignacio Cases, and Andrei Barbu.
\newblock Brain{BERT}: Self-supervised representation learning for intracranial recordings.
\newblock In {\em The Eleventh International Conference on Learning Representations}, February 2023.

\bibitem{chen2024eegformer}
Yuqi Chen, Kan Ren, Kaitao Song, Yansen Wang, Yifan Wang, Dongsheng Li, and Lili Qiu.
\newblock {EEGF}ormer: Towards transferable and interpretable large-scale {EEG} foundation model.
\newblock In {\em AAAI 2024 Spring Symposium on Clinical Foundation Models}, February 2024.

\bibitem{jiang2024large}
Weibang Jiang, Liming Zhao, and Bao liang Lu.
\newblock Large brain model for learning generic representations with tremendous {EEG} data in {BCI}.
\newblock In {\em The Twelfth International Conference on Learning Representations}, January 2024.

\bibitem{wang2024cbramod}
Jiquan Wang, Sha Zhao, Zhiling Luo, Yangxuan Zhou, Haiteng Jiang, Shijian Li, Tao Li, and Gang Pan.
\newblock {CB}ra{M}od: A criss-cross brain foundation model for {EEG} decoding.
\newblock In {\em The Thirteenth International Conference on Learning Representations}, January 2025.

\bibitem{zhang2023brant}
Daoze Zhang, Zhizhang Yuan, Yang Yang, Junru Chen, Jingjing Wang, and Yafeng Li.
\newblock Brant: Foundation model for intracranial neural signal.
\newblock In {\em Thirty-seventh Conference on Neural Information Processing Systems}, September 2023.

\bibitem{tang2021self}
Siyi Tang, Jared Dunnmon, Khaled~Kamal Saab, Xuan Zhang, Qianying Huang, Florian Dubost, Daniel Rubin, and Christopher Lee-Messer.
\newblock Self-supervised graph neural networks for improved electroencephalographic seizure analysis.
\newblock In {\em International Conference on Learning Representations}, January 2022.

\bibitem{dimofte2025cerebro}
Alexandru Dimofte, Glenn~Anta Bucagu, Thorir~Mar Ingolfsson, Xiaying Wang, Andrea Cossettini, Luca Benini, and Yawei Li.
\newblock {CER}e{B}r{O}: Compact encoder for representations of brain oscillations using efficient alternating attention, January 2025.

\bibitem{chau2024population}
Geeling Chau, Christopher Wang, Sabera~J Talukder, Vighnesh Subramaniam, Saraswati Soedarmadji, Yisong Yue, Boris Katz, and Andrei Barbu.
\newblock Population transformer: Learning population-level representations of neural activity.
\newblock In {\em The Thirteenth International Conference on Learning Representations}, January 2025.

\bibitem{saeed2021learning}
Aaqib Saeed, David Grangier, Olivier Pietquin, and Neil Zeghidour.
\newblock Learning from heterogeneous eeg signals with differentiable channel reordering.
\newblock In {\em ICASSP 2021-2021 IEEE international conference on acoustics, speech and signal processing (ICASSP)}, pages 1255--1259. IEEE, 2021.

\bibitem{lee2019set}
Juho Lee, Yoonho Lee, Jungtaek Kim, Adam Kosiorek, Seungjin Choi, and Yee~Whye Teh.
\newblock Set transformer: A framework for attention-based permutation-invariant neural networks.
\newblock In {\em International conference on machine learning}, pages 3744--3753. PMLR, June 2019.

\bibitem{jaegle2021perceiver}
Andrew Jaegle, Sebastian Borgeaud, Jean-Baptiste Alayrac, Carl Doersch, Catalin Ionescu, David Ding, Skanda Koppula, Daniel Zoran, Andrew Brock, Evan Shelhamer, Olivier~J Henaff, Matthew Botvinick, Andrew Zisserman, Oriol Vinyals, and Joao Carreira.
\newblock Perceiver {IO}: A general architecture for structured inputs \& outputs.
\newblock In {\em International Conference on Learning Representations}, January 2022.

\bibitem{wu2018group}
Yuxin Wu and Kaiming He.
\newblock Group normalization.
\newblock {\em International Journal of Computer Vision}, 128(3):742--755, July 2019.

\bibitem{hendrycks2016gaussian}
Dan Hendrycks and Kevin Gimpel.
\newblock Gaussian error linear units (gelus), June 2016.

\bibitem{mildenhallNerf}
Ben Mildenhall, Pratul~P. Srinivasan, Matthew Tancik, Jonathan~T. Barron, Ravi Ramamoorthi, and Ren Ng.
\newblock {NeRF}: representing scenes as neural radiance fields for view synthesis.
\newblock {\em Commun. ACM}, 65(1):99–106, December 2021.

\bibitem{su2024roformer}
Jianlin Su, Murtadha Ahmed, Yu~Lu, Shengfeng Pan, Wen Bo, and Yunfeng Liu.
\newblock Ro{F}ormer: Enhanced transformer with rotary position embedding.
\newblock {\em Neurocomputing}, 568:127063, February 2024.

\bibitem{ingolfsson2022}
Thorir~Mar Ingolfsson, Andrea Cossettini, Simone Benatti, and Luca Benini.
\newblock Energy-efficient tree-based eeg artifact detection.
\newblock In {\em 2022 44th Annual International Conference of the IEEE Engineering in Medicine \& Biology Society (EMBC)}, pages 3723--3728. IEEE, 2022.

\bibitem{jing2023development}
Jin Jing, Wendong Ge, Shenda Hong, Marta~Bento Fernandes, Zhen Lin, Chaoqi Yang, Sungtae An, Aaron~F. Struck, Aline Herlopian, Ioannis Karakis, Jonathan~J. Halford, Marcus~C. Ng, Emily~L. Johnson, Brian~L. Appavu, Rani~A. Sarkis, Gamaleldin Osman, Peter~W. Kaplan, Monica~B. Dhakar, Lakshman~Arcot Jayagopal, Zubeda Sheikh, Olga Taraschenko, Sarah Schmitt, Hiba~A. Haider, Jennifer~A. Kim, Christa~B. Swisher, Nicolas Gaspard, Mackenzie~C. Cervenka, Andres A.~Rodriguez Ruiz, Jong~Woo Lee, Mohammad Tabaeizadeh, Emily~J. Gilmore, Kristy Nordstrom, Ji~Yeoun Yoo, Manisha~G. Holmes, Susan~T. Herman, Jennifer~A. Williams, Jay Pathmanathan, Fábio~A. Nascimento, Ziwei Fan, Samaneh Nasiri, Mouhsin~M. Shafi, Sydney~S. Cash, Daniel~B. Hoch, Andrew~J. Cole, Eric~S. Rosenthal, Sahar~F. Zafar, Jimeng Sun, and M.~Brandon Westover.
\newblock Development of expert-level classification of seizures and rhythmic and periodic patterns during {EEG} interpretation.
\newblock {\em Neurology}, 100(17):e1750--e1762, April 2023.

\bibitem{yang2021self}
Chaoqi Yang, Cao Xiao, M.~Brandon Westover, and Jimeng Sun.
\newblock Self-supervised electroencephalogram representation learning for automatic sleep staging: Model development and evaluation study.
\newblock {\em JMIR AI}, 2:e46769, July 2021.

\bibitem{peh2022transformer}
Wei~Yan Peh, Yuanyuan Yao, and Justin Dauwels.
\newblock Transformer convolutional neural networks for automated artifact detection in scalp {EEG}, July 2022.

\bibitem{li2022motor}
Hongli Li, Man Ding, Ronghua Zhang, and Chunbo Xiu.
\newblock Motor imagery {EEG} classification algorithm based on cnn-lstm feature fusion network.
\newblock {\em Biomedical Signal Processing and Control}, 72:103342, February 2022.

\bibitem{song2021transformer}
Yonghao Song, Xueyu Jia, Lie Yang, and Longhan Xie.
\newblock Transformer-based spatial-temporal feature learning for {EEG} decoding, June 2021.

\bibitem{mohammadi2024eeg2rep}
Navid~Mohammadi Foumani, Geoffrey Mackellar, Soheila Ghane, Saad Irtza, Nam Nguyen, and Mahsa Salehi.
\newblock Eeg2rep: Enhancing self-supervised {EEG} representation through informative masked inputs.
\newblock In {\em Proceedings of the 30th ACM SIGKDD Conference on Knowledge Discovery and Data Mining}, KDD ’24, pages 5544--5555. ACM, August 2024.

\bibitem{tegon2025femba}
Anna Tegon, Thorir~Mar Ingolfsson, Xiaying Wang, Luca Benini, and Yawei Li.
\newblock {FEMBA}: Efficient and scalable {EEG} analysis with a bidirectional mamba foundation model, February 2025.

\bibitem{lawhern2018eegnet}
Vernon~J Lawhern, Amelia~J Solon, Nicholas~R Waytowich, Stephen~M Gordon, Chou~P Hung, and Brent~J Lance.
\newblock Eegnet: a compact convolutional neural network for eeg-based brain–computer interfaces.
\newblock {\em Journal of Neural Engineering}, 15(5):056013, July 2018.

\bibitem{tang2023modeling}
Siyi Tang, Jared~A Dunnmon, Qu~Liangqiong, Khaled~K Saab, Tina Baykaner, Christopher Lee-Messer, and Daniel~L Rubin.
\newblock Modeling multivariate biosignals with graph neural networks and structured state space models.
\newblock In Bobak~J. Mortazavi, Tasmie Sarker, Andrew Beam, and Joyce~C. Ho, editors, {\em Proceedings of the Conference on Health, Inference, and Learning}, volume 209 of {\em Proceedings of Machine Learning Research}, pages 50--71. PMLR, June 2023.

\end{thebibliography}

\clearpage
\section*{NeurIPS Paper Checklist}

\begin{enumerate}

\item {\bf Claims}
    \item[] Question: Do the main claims made in the abstract and introduction accurately reflect the paper's contributions and scope?
    \item[] Answer: \answerYes{} 
    \item[] Justification: We propose a topology-agnostic foundation model with a focus on computational efficiency. We tediously compare our model based on downstream task performance, including datasets with different electrode topologies, and computational efficiency with respect to prior art in Section \ref{sec:results}.
    

\item {\bf Limitations}
    \item[] Question: Does the paper discuss the limitations of the work performed by the authors?
    \item[] Answer: \answerYes{} 
    \item[] Justification: We discuss the limitations in Section \ref{subsec:downstream_results} and \ref{sec:conclusion}. These limitations include a lower performance on the SEED-V dataset, which is an unseen and denser topology compared to pre-training datasets, and the reliance on positional encodings learned during pre-training.  

\item {\bf Theory assumptions and proofs}
    \item[] Question: For each theoretical result, does the paper provide the full set of assumptions and a complete (and correct) proof?
    \item[] Answer: \answerNA{} 
    \item[] Justification: The paper doesn't include theoretical results. We discuss the computational complexity of different methods theoretically, but report the empirical computational cost in Figure \ref{fig:scaling_plots_combined}.

    \item {\bf Experimental result reproducibility}
    \item[] Question: Does the paper fully disclose all the information needed to reproduce the main experimental results of the paper to the extent that it affects the main claims and/or conclusions of the paper (regardless of whether the code and data are provided or not)?
    \item[] Answer: \answerYes{} 
    \item[] Justification: We thoroughly explain our model architecture and experimental details in Section \ref{sec:method} and \ref{subsec:setup}. We also use public datasets and describe the preprocessing details. We will also release the code and the pre-trained model weights upon publication. 

\item {\bf Open access to data and code}
    \item[] Question: Does the paper provide open access to the data and code, with sufficient instructions to faithfully reproduce the main experimental results, as described in supplemental material?
    \item[] Answer: \answerYes{} 
    \item[] Justification: We will publish the code and the pre-trained model weights upon publication. 

\item {\bf Experimental setting/details}
    \item[] Question: Does the paper specify all the training and test details (e.g., data splits, hyperparameters, how they were chosen, type of optimizer, etc.) necessary to understand the results?
    \item[] Answer: \answerYes{} 
    \item[] Justification: The details are discussed in detail in Section \ref{subsec:setup} as well as in \ref{app:model_details} in the Appendix.

\item {\bf Experiment statistical significance}
    \item[] Question: Does the paper report error bars suitably and correctly defined or other appropriate information about the statistical significance of the experiments?
    \item[] Answer: \answerYes{} 
    \item[] Justification: We report the mean and the standard deviation of results obtained from initializing the models with different seeds, as described in Section \ref{subsec:setup}.

\item {\bf Experiments compute resources}
    \item[] Question: For each experiment, does the paper provide sufficient information on the computer resources (type of compute workers, memory, time of execution) needed to reproduce the experiments?
    \item[] Answer: \answerYes{} 
    \item[] Justification: We report the GPU types and the training duration in Section \ref{subsec:setup} and \ref{app:exp_settings}.
    
\item {\bf Code of ethics}
    \item[] Question: Does the research conducted in the paper conform, in every respect, with the NeurIPS Code of Ethics \url{https://neurips.cc/public/EthicsGuidelines}?
    \item[] Answer: \answerYes{} 
    \item[] Justification: We use public EEG datasets that conform to ethical guidelines and are commonly used in literature. 

\item {\bf Broader impacts}
    \item[] Question: Does the paper discuss both potential positive societal impacts and negative societal impacts of the work performed?
    \item[] Answer: \answerYes{} 
    \item[] Justification: The societal impacts are discussed in Section \ref{sec:conclusion}.
    
\item {\bf Safeguards}
    \item[] Question: Does the paper describe safeguards that have been put in place for responsible release of data or models that have a high risk for misuse (e.g., pretrained language models, image generators, or scraped datasets)?
    \item[] Answer: \answerNA{} 
    \item[] Justification: The paper does not introduce any such risk. 

\item {\bf Licenses for existing assets}
    \item[] Question: Are the creators or original owners of assets (e.g., code, data, models), used in the paper, properly credited and are the license and terms of use explicitly mentioned and properly respected?
    \item[] Answer: \answerYes{} 
    \item[] Justification: We correctly cite and conform to the licenses of the papers that produced the public datasets we used. 

\item {\bf New assets}
    \item[] Question: Are new assets introduced in the paper well documented and is the documentation provided alongside the assets?
    \item[] Answer: \answerNA{} 
    \item[] Justification: We do not release any new assets yet. The code and the released model will be released open source with the appropriate license and documentation.

\item {\bf Crowdsourcing and research with human subjects}
    \item[] Question: For crowdsourcing experiments and research with human subjects, does the paper include the full text of instructions given to participants and screenshots, if applicable, as well as details about compensation (if any)? 
    \item[] Answer: \answerNA{} 
    \item[] Justification: The paper does not include research with human subjects. We use existing public datasets in the area.

\item {\bf Institutional review board (IRB) approvals or equivalent for research with human subjects}
    \item[] Question: Does the paper describe potential risks incurred by study participants, whether such risks were disclosed to the subjects, and whether Institutional Review Board (IRB) approvals (or an equivalent approval/review based on the requirements of your country or institution) were obtained?
    \item[] Answer: \answerNA{} 
    \item[] Justification: The paper does not include research with human subjects. We use existing public datasets in the area.

\item {\bf Declaration of LLM usage}
    \item[] Question: Does the paper describe the usage of LLMs if it is an important, original, or non-standard component of the core methods in this research? Note that if the LLM is used only for writing, editing, or formatting purposes and does not impact the core methodology, scientific rigorousness, or originality of the research, declaration is not required.
    \item[] Answer: \answerNA{} 
    \item[] Justification: LLMs were only used for writing, editing, amd formatting the content.

\end{enumerate}
\newpage
\appendix
\section{Appendix}
\label{sec:appendix}

This appendix provides supplementary details that complement the main paper: additional efficiency analyses, reporting notes for statistical testing, small implementation clarifications, and extra qualitative comparisons.

\subsection{Model Architecture Details}
\label{app:model_details}

The following tables show the hyperparameter setup for the pre-training and the downstream fine-tuning for LUNA.

\subsubsection{Hyperparameters for pre-training}
\begin{table}[htbp!]
\centering
\caption{Hyperparameters for EEG pre-training.}
\begin{tabular}{@{}ccccc@{}}
\toprule
\multicolumn{2}{c}{\textbf{Hyperparameters}} & \textbf{LUNA-Base} & \textbf{LUNA-Large} & \textbf{LUNA-Huge} \cr
\midrule
\multirow{5}*{Temporal Encoder} & Input channels & \{1,8,8\} & \{1,16,16\} & \{1,32,32\} \\
& Output channels & \{16,16,16\} & \{24,24,24\} & \{32,32,32\}  \\
& Kernel size & \multicolumn{3}{c}{\{20,3,3\}} \\
& Stride & \multicolumn{3}{c}{\{10,1,1\}} \\
& Padding & \multicolumn{3}{c}{\{9,1,1\}} \\
\midrule
\multicolumn{2}{c}{Patch size} & \multicolumn{3}{c}{40} \\
\multicolumn{2}{c}{Transformer encoder layers} & 8 & 10 & 24 \\
\multicolumn{2}{c}{Number of queries} & 4 & 6 & 8 \\
\multicolumn{2}{c}{Query size} & 64 & 96 & 128 \\
\multicolumn{2}{c}{Hidden size} & 256 & 576 & 1024\\
\multicolumn{2}{c}{MLP size} & 1024 & 2304 & 4096 \\
\multicolumn{2}{c}{Attention head number} & 8 & 12 & 16 \\
\midrule
\multicolumn{2}{c}{Batch size per GPU} & 2040 & 2040 & 720 \\
\multicolumn{2}{c}{Total batch size} & 8160 & 8160 & 11520 \\
\multicolumn{2}{c}{Peak learning rate} & \multicolumn{3}{c}{1.25e-4}\\
\multicolumn{2}{c}{Minimal learning rate} & \multicolumn{3}{c}{2.5e-7} \\
\multicolumn{2}{c}{Learning rate scheduler} & \multicolumn{3}{c}{Cosine} \\
\multicolumn{2}{c}{Optimizer} & \multicolumn{3}{c}{AdamW} \\
\multicolumn{2}{c}{Adam $\beta$} & \multicolumn{3}{c}{(0.9,0.98)} \\
\multicolumn{2}{c}{Weight decay} & \multicolumn{3}{c}{0.05} \\
\multicolumn{2}{c}{Total epochs} & \multicolumn{3}{c}{60} \\
\multicolumn{2}{c}{Warmup epochs} & \multicolumn{3}{c}{10} \\
\multicolumn{2}{c}{Loss type} & \multicolumn{3}{c}{Smooth-L1} \\
\multicolumn{2}{c}{Non-masked region loss coefficient} & \multicolumn{3}{c}{0.05} \\
\multicolumn{2}{c}{Query specialization loss coefficient} & \multicolumn{3}{c}{0.8} \\
\midrule
\multicolumn{2}{c}{Gradient clipping} & \multicolumn{3}{c}{1} \\
\multicolumn{2}{c}{Mask ratio} & \multicolumn{3}{c}{0.5} \\
\multicolumn{2}{c}{Precision} & \multicolumn{3}{c}{bf16-mixed} \\

\bottomrule
\end{tabular}
\label{tab:pre-training}
\end{table}

\clearpage
\subsubsection{Hyperparameters for downstream fine-tuning}

\begin{table}[htbp!]
\centering
\caption{Hyperparameters for downstream fine-tuning.}
\begin{tabular}{@{}cc@{}}
\toprule
\textbf{Hyperparameters} & \textbf{Values} \cr
\midrule
Batch size per GPU & 512 \\
Peak learning rate & 1e-4 \\
Minimal learning rate & 5e-6 \\
Learning rate scheduler & Cosine \\
Optimizer & AdamW \\
Adam $\beta$ & (0.9,0.999) \\
Weight decay & 0.05 \\
Total epochs & 50 \\
Early stopping patience & 10 \\
Warmup epochs & 5 \\
Drop path & 0.1 (B/L) 0.2 (H) \\
Layer-wise learning rate decay & 0.5 (B) 0.8 (L/H) \\
Label smoothing (multi-class classification) & 0.1 \\
\bottomrule
\end{tabular}
\label{tab:finetuning}
\end{table}

\subsubsection{Complexity Analysis}
\label{app:biot_scaling}
The computational complexity of key attention stages and a comparison with alternatives are shown in \ref{tab:app_complexity_breakdown} and 
\ref{tab:app_attention_complexity}.

\begin{table}[htbp!]
\centering
\caption{Complexity Breakdown of LUNA Encoder Stages.}
\label{tab:app_complexity_breakdown}
\begin{tabular}{@{}lcc@{}}
\toprule
\textbf{Stage} & \textbf{Input Shape} & \textbf{Complexity} \\
\midrule
Channel-Unification Module (Cross-Attn) & $(B \cdot S) \times C \times E$ & $O(B \cdot S \cdot Q \cdot C \cdot E)$ \\
Query Self-Attention & $(B \cdot S) \times Q \times E$ & $O(B \cdot S \cdot Q^2 \cdot E)$ \\
Patch-wise Attention Encoder (Self-Attn) & $B \times S \times (Q\cdot E)$ & $O(B \cdot S^2 \cdot Q \cdot E)$ \\ 
\bottomrule
\end{tabular}
\end{table}

\begin{table}[htbp!]
\centering
\caption{Attention Complexity Comparison.}
\label{tab:app_attention_complexity}
\begin{tabular}{@{}lc@{}}
\toprule
\textbf{Method} & \textbf{Bottleneck Complexity} \\ 
\midrule
LUNA (Latent Space Attention) & $O(B \cdot S^2 \cdot Q \cdot E)$ \textit{or} $O(B \cdot S \cdot Q \cdot C \cdot E)$ \\
Full-Attention (e.g., LaBraM) & $O(B \cdot S^2 \cdot C^2 \cdot E)$ \\
Alternating Attention (Patches, e.g., CBraMod) & $O(B \cdot S^2 \cdot C \cdot E)$ \\
Alternating Attention (Channels, e.g., CBraMod) & $O(B \cdot S \cdot C^2 \cdot E)$ \\
\bottomrule
\end{tabular}
\end{table}

\paragraph{BIOT vs.\ LUNA scaling.}
Tables~\ref{tab:biot_patches}--\ref{tab:biot_channels} report GFLOPs and peak activation memory across varying patch and channel counts.

\begin{table}[h]
\centering
\caption{Scaling with patch count (GFLOPs and MiB per forward).}
\label{tab:biot_patches}
\small
\begin{tabular}{lrrrr}
\toprule
Model & \#Patches & FLOPs (G) & Memory (MiB) \\
\midrule
BIOT & 2000 & 1143 & 3534 \\
LUNA-Base & 2000 & 253 & 1835 \\
LUNA-Large & 2000 & 1073 & 2969 \\
LUNA-Huge & 2000 & 6570 & 5549 \\
BIOT & 3000 & 1714 & 5231 \\
LUNA-Base & 3000 & 478 & 3873 \\
LUNA-Large & 3000 & 1886 & 6041 \\
LUNA-Huge & 3000 & 11035 & 9685 \\
BIOT & 4000 & 2286 & 6931 \\
LUNA-Base & 4000 & 768 & 6678 \\
LUNA-Large & 4000 & 2884 & 10265 \\
LUNA-Huge & 4000 & 16286 & 15344 \\
\bottomrule
\end{tabular}
\end{table}

\begin{table}[h]
\centering
\caption{Scaling with channel count (GFLOPs and MiB per forward).}
\label{tab:biot_channels}
\small
\begin{tabular}{lrrrr}
\toprule
Model & \#Channels & FLOPs (G) & Memory (MiB) \\
\midrule
BIOT & 6000 & 3117 & 9270 \\
LUNA-Base & 6000 & 18 & 1571 \\
LUNA-Large & 6000 & 43 & 2410 \\
LUNA-Huge & 6000 & 112 & 4198 \\
BIOT & 7000 & 3637 & 10791 \\
LUNA-Base & 7000 & 20 & 1826 \\
LUNA-Large & 7000 & 49 & 2777 \\
LUNA-Huge & 7000 & 122 & 4680 \\
BIOT & 8000 & 4156 & 12307 \\
LUNA-Base & 8000 & 23 & 2069 \\
LUNA-Large & 8000 & 55 & 3139 \\
LUNA-Huge & 8000 & 133 & 5163 \\
\bottomrule
\end{tabular}
\end{table}

\subsection{Dataset and Preprocessing Details}
\label{app:datasets}

\paragraph{Datasets Used}
We use publicly available EEG datasets,
provided in \ref{tab:app_dataset_summary}.

\begin{table}[htbp!]
\centering
\caption{Summary of Datasets Used.}
\label{tab:app_dataset_summary}
\resizebox{\columnwidth}{!} {
\begin{tabular}{@{}lccccc@{}}
\toprule
\textbf{Dataset} & \textbf{\# Subjects} & \textbf{\# Samples (Train/Val/Test or Total)} & \textbf{Hours of Recordings} & \textbf{\# Channels} & \textbf{Montage Used} \\
\midrule
TUEG (Pre-train) & 14,987 & 15,686,874 (Total) & 21,787.32 & 20 or 22 & Bipolar \\
Siena (Pre-train) & 14 & 101,520 (Total) & 141.0 & 29 & Unipolar \\
TUAB & 2,329 & 591,357 / 154,938 / 74,010 & 1,139.31 & 22 & Bipolar \\
TUAR & 213 & 49,241 / 5,870 / 5,179 & 83.74 & 22 & Bipolar \\
TUSL & 38 & 16,088 / 1,203 / 2,540 & 27.54 & 22 & Bipolar \\
SEED-V & 15 & 43,328 / 43,360 / 31,056 & 32.70 & 62 & Unipolar \\
\bottomrule
\end{tabular}}
\end{table}

\subsection{Experimental Settings}
\label{app:exp_settings}

\paragraph{Pre-training}
LUNA is pre-trained using a masked patch reconstruction task. Key hyperparameters are listed in \ref{tab:pre-training}.

\paragraph{Computational Resources}

Experiments were conducted using NVIDIA A100 GPUs. Pre-training took approximately 1 day on 8 GPUs for the base and large models and 16 GPUs for the huge model.

\subsection{Additional Quantitative Results}
\label{app:quant_extra}

\paragraph{Training Curves}
The pre-training loss curves for LUNA-Base are shown in \ref{fig:app_loss_curve}. The reconstruction loss drops shows and initial plateau then drops slowly over the epochs, while the query specialization shows a jump and then a slow decrease, indicating more orthogonal query usage over time. The initial drop of the query specialization might be due to a trivial case where a query attends to only one channel. The queries learn to attend to their own specialized areas afterwards while covering all the channels in the input.

\begin{figure}[htbp!]
    \centering
    \includegraphics[width=0.8\linewidth]{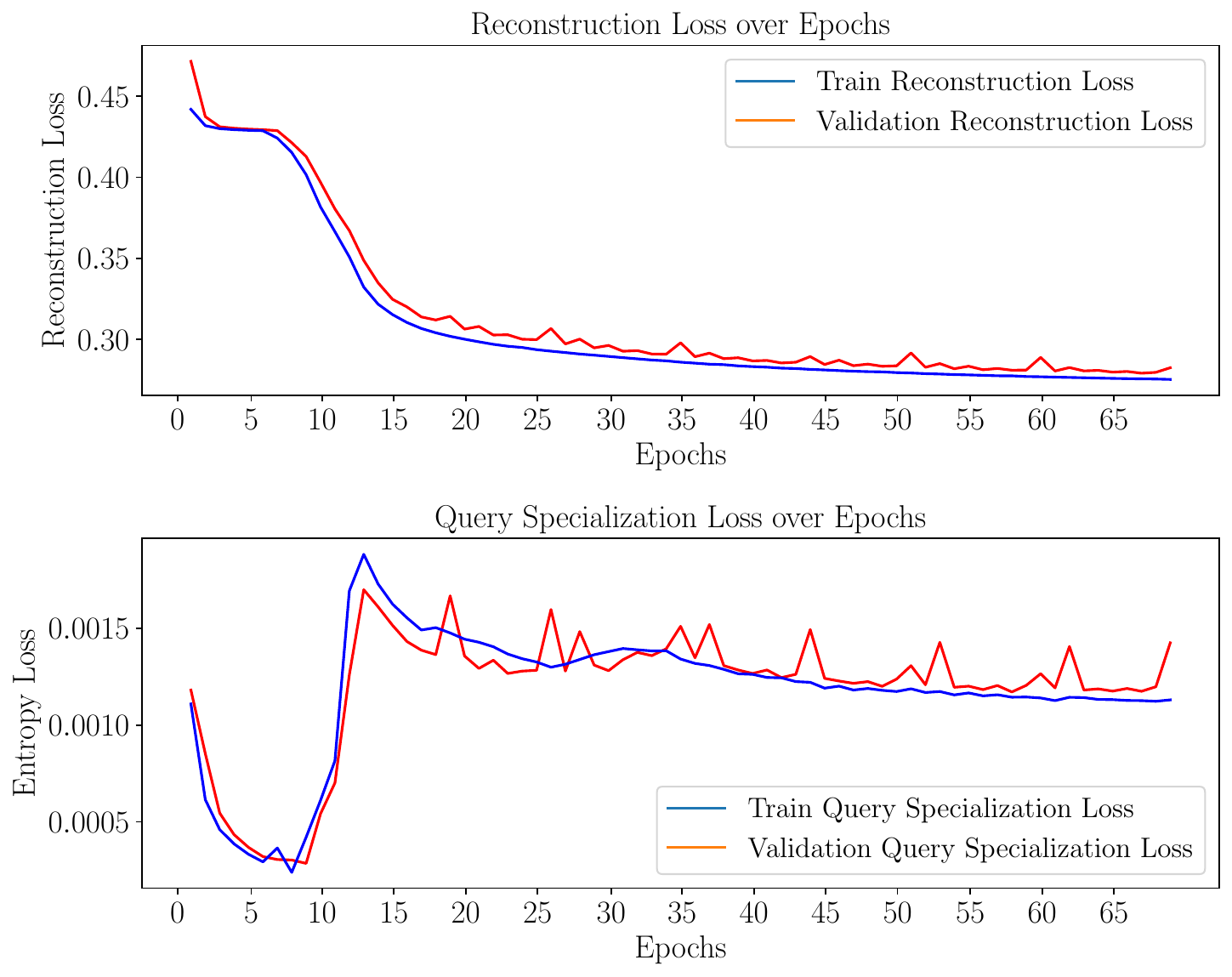} 
    \caption{Loss curves during pre-training for LUNA-Base (Reconstruction and Query Specialization Loss).}
    \label{fig:app_loss_curve}
\end{figure}

\subsection{Additional Visualizations}
\label{app:viz_extra}

\paragraph{Reconstruction Examples} 
Figures
\ref{fig:app_20_channels}, \ref{fig:app_22_channels}, \ref{fig:app_29_channels} show examples of the model reconstructing masked patches (gray regions) for inputs with 20, 22, and 29 channels, respectively. The reconstructions capture the underlying signal trend and demonstrate robustness across different topologies seen during pre-training.

\begin{figure}[htbp!]
    \centering
\includegraphics[height=0.25\textheight]{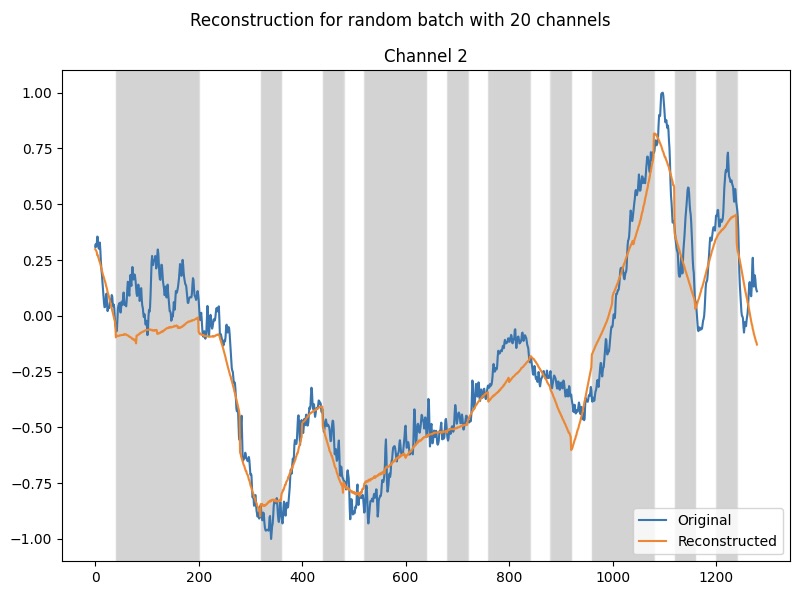}
    \caption{Example reconstruction on input with 20 channels (masked regions in gray).}
    \label{fig:app_20_channels}
\end{figure}
\begin{figure}[htbp!]
    \centering
    \includegraphics[height=0.25\textheight]{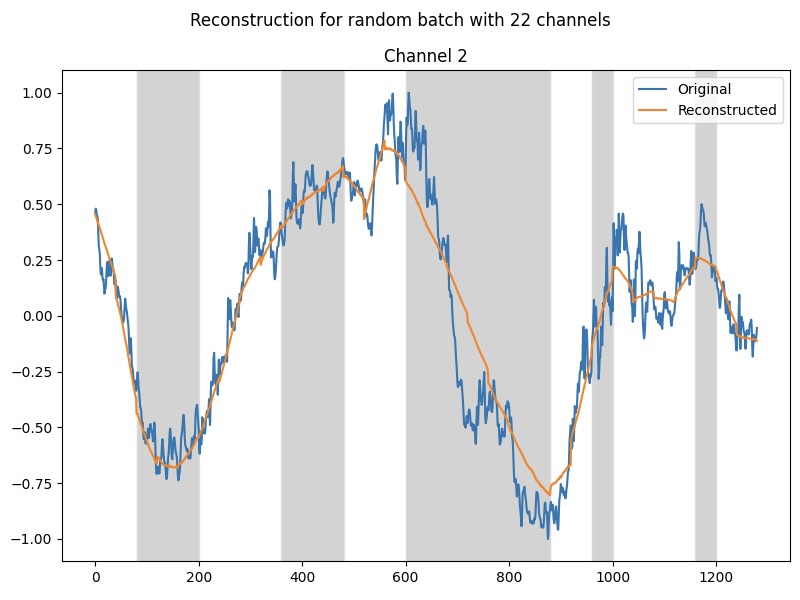}
    \caption{Example reconstruction on input with 22 channels (masked regions in gray).}
    \label{fig:app_22_channels}
\end{figure}
\begin{figure}[htbp!]
    \centering
    \includegraphics[height=0.25\textheight]{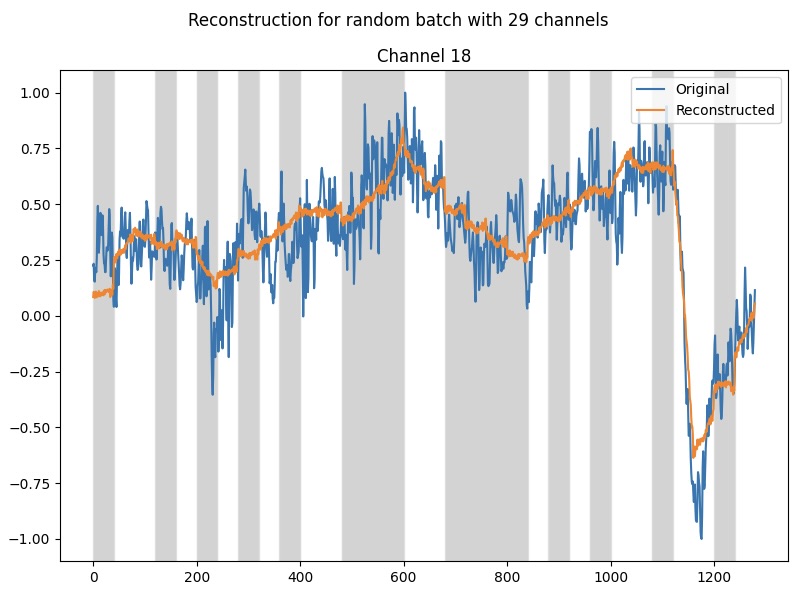}
    \caption{Example reconstruction on input with 29 channels (masked regions in gray).}
    \label{fig:app_29_channels}
\end{figure}

\subsection{Trade-off between $Q$ and $E$}
\label{app:q_tradeoff}
\begin{table}[h]
\centering
\caption{LUNA-Base variants with fixed $Q\cdot E{=}256$.}
\resizebox{\columnwidth}{!} {
\begin{tabular}{lrrrrrrr}
\toprule
Variant ($Q\times E$) & $Q$ & $E$ & TUAB AUROC & TUAB AUPRC & TUAR AUROC & TUAR AUPRC & TUSL AUROC \\
\midrule
$4\times 64$   & 4  & 64  & 0.887 & 0.895 & 0.902 & 0.495 & 0.767 \\
$2\times 128$  & 2  & 128 & 0.885 & 0.890 & 0.885 & 0.501 & 0.759 \\
$8\times 32$   & 8  & 32  & 0.884 & 0.892 & 0.899 & 0.505 & 0.766 \\
$16\times 16$  & 16 & 16  & 0.874 & 0.881 & 0.866 & 0.487 & 0.757 \\
\bottomrule
\end{tabular}}
\end{table}

\noindent\textit{Observation.} Increasing $Q$ at the expense of $E$ degrades performance; too few queries can also bottleneck capacity. A balanced $Q{\times}E$ (e.g., $4{\times}64$) works well under the same latent budget.

\subsection{Bipolar montage electrode pairs}
\label{app:bipolar_pairs}
We use the following longitudinal pairs to construct the bipolar montage for TUEG, TUAR, TUSL, and TUAB (left/right symmetric sets):
\begin{itemize}\itemsep2pt
  \item Fp1--F7, F7--T3, T3--T5, T5--O1,\quad Fp2--F8, F8--T4, T4--T6, T6--O2 T3--C3, C3--CZ
  \item Fp1--F3, F3--C3, C3--P3, P3--O1,\quad Fp2--F4, F4--C4, C4--P4, P4--O2 CZ--C4, C4--T4
\end{itemize}

\subsection{Edge deployment reference}
Typical low-power edge SoCs used in wearables/IoT offer single-digit to few dozen MB of RAM and on-chip/SiP compute in the 10--100 GOPS range. Under these constraints, LUNA-Base ($\sim$7M params; $\sim$14\,MB at 16-bit) and its measured GFLOPs per window (Fig~\ref{fig:channel_scaling}) fit comfortably within real-time budgets, whereas quadratic-in-$C$ spatial attention and larger activation footprints in some baselines make deployment more challenging at higher channel counts.

\subsection{Significance Testing}
\label{app:sig_tests}
Unless noted, we report mean~$\pm$~s.d.\ over matched seeds. For ablations we ran two-sided paired $t$-tests across seeds for specific comparisons requested by reviewers. For the TUAR AUROC comparison of the model with vs.\ without the query specialization loss, the paired $t$-test yielded $p{=}0.2136$, i.e., not statistically significant at $\alpha{=}0.05$. Observed AUROC deltas across ablations were small (absolute $\le 0.01$).

\begin{table}[h]
\centering
\caption{Paired $t$-test on TUAR AUROC for the specialization-loss ablation.}
\label{tab:sig_tests}
\small
\begin{tabular}{lcc}
\toprule
Comparison & Mean $\Delta$ (w/o $-$ full) & $p$ \\
\midrule
w/o specialization vs.\ full & $-0.007$ & $0.2136$ \\
\bottomrule
\end{tabular}
\end{table}

\paragraph{Practical tolerance.}
We adopt a pragmatic tolerance of $\pm 0.01$ AUROC for considering two variants practically equivalent on these datasets; all reported ablations fall within this band.

\subsection{Effect of Specialization Loss on Query Maps}
\label{app:spec_loss_viz}
Figure~\ref{fig:spec_loss_compare} depicts spatial attention maps of the $Q$ queries when the specialization loss is \emph{removed}. In this setting, two queries converge to coarse lateralized patterns (left and right longitudinal chains), while the remaining queries display broad, overlapping support over fronto–central and midline sites with weaker focal peaks, indicating partial redundancy and gaps in complementary coverage. Overall, the maps exhibit higher overlap and reduced distinctiveness across queries compared to the model trained with the loss, where query maps are more complementary and less overlapping Fig.~\ref{fig:query_analysis}).

\begin{figure}[h]
\centering
\includegraphics[width=0.95\linewidth]{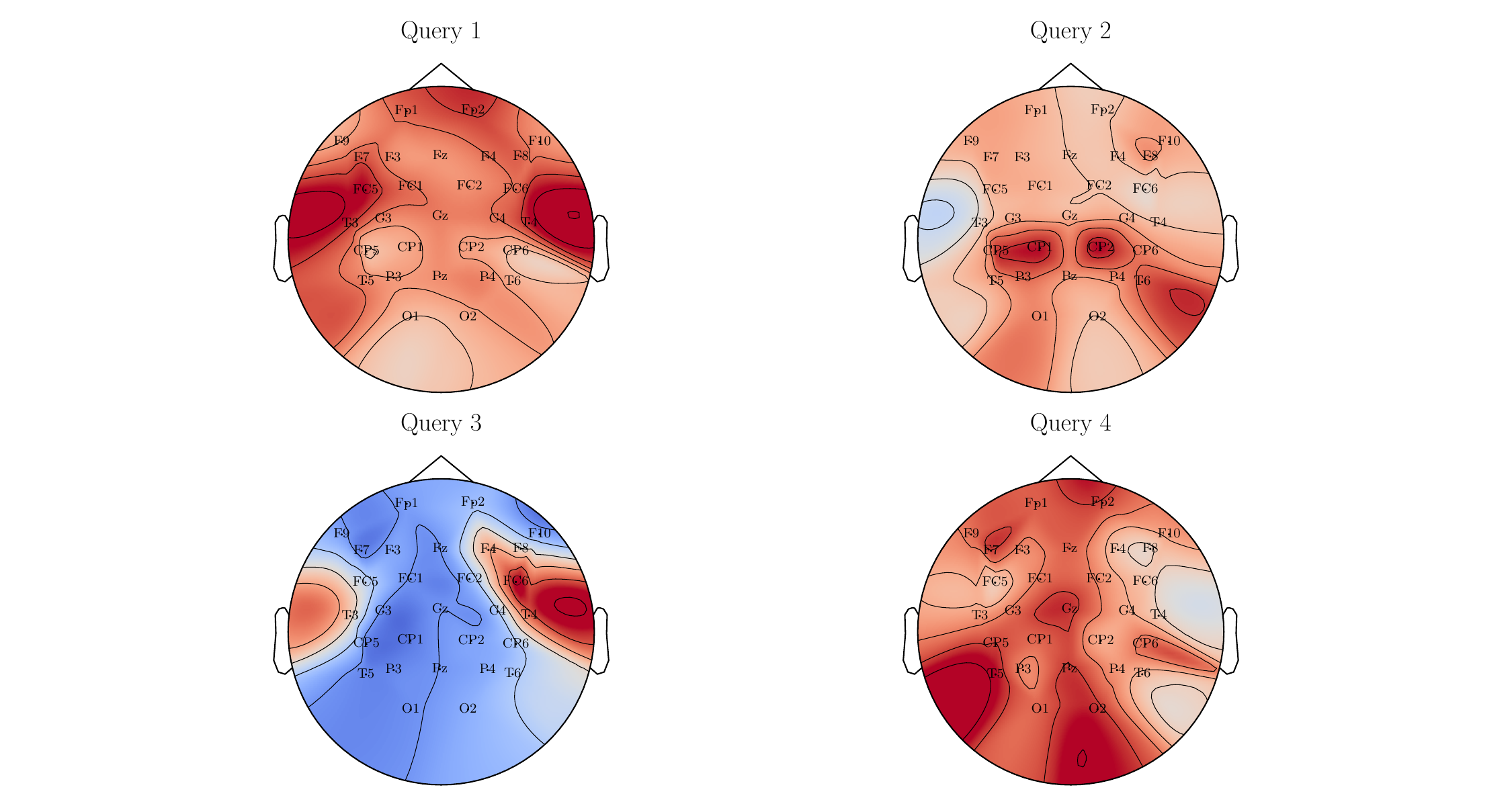}
\caption{Query attention visualization without specialization loss.}
\label{fig:spec_loss_compare}
\end{figure}

\subsection{t-SNE of Raw Frequency Features vs.\ LUNA Features}
We compute per-segment frequency features (magnitude/phase statistics per band, averaged across channels) and compare 2D t-SNE embeddings to those of LUNA’s latent features. Raw features exhibit less separation beyond clear artifacts on TUAR; LUNA features show tighter clustering aligned with labels.
\begin{figure}[htbp!]
    \centering
    \begin{subfigure}[b]{0.48\textwidth}
        \centering
        \includegraphics[width=\linewidth]{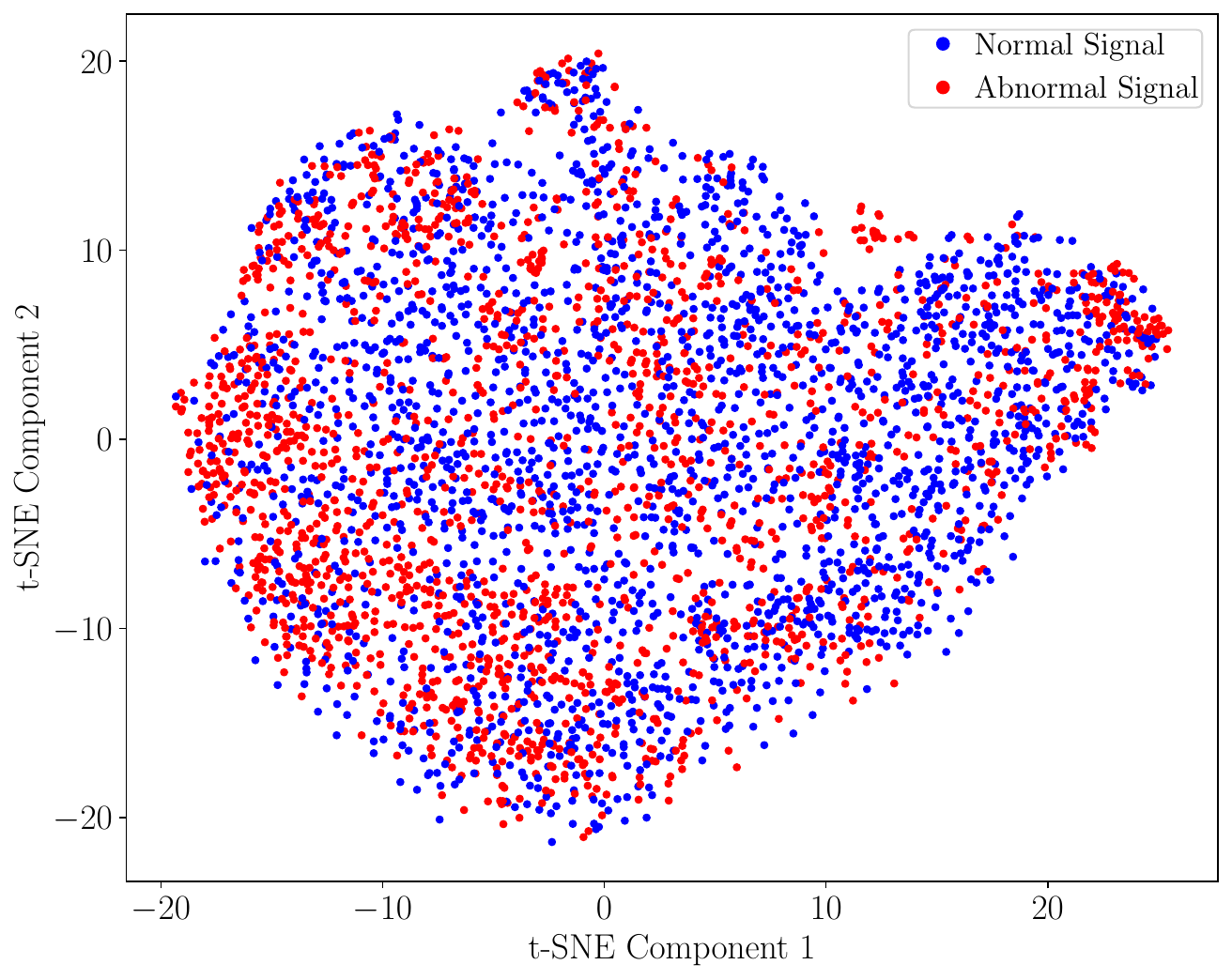}
        \caption{TUAB dataset (Normal vs. Abnormal Signal).}
        \label{fig:app_TUAB_emb_raw}
    \end{subfigure}
    \hfill 
    \begin{subfigure}[b]{0.48\textwidth}
        \centering
        \includegraphics[width=\linewidth]{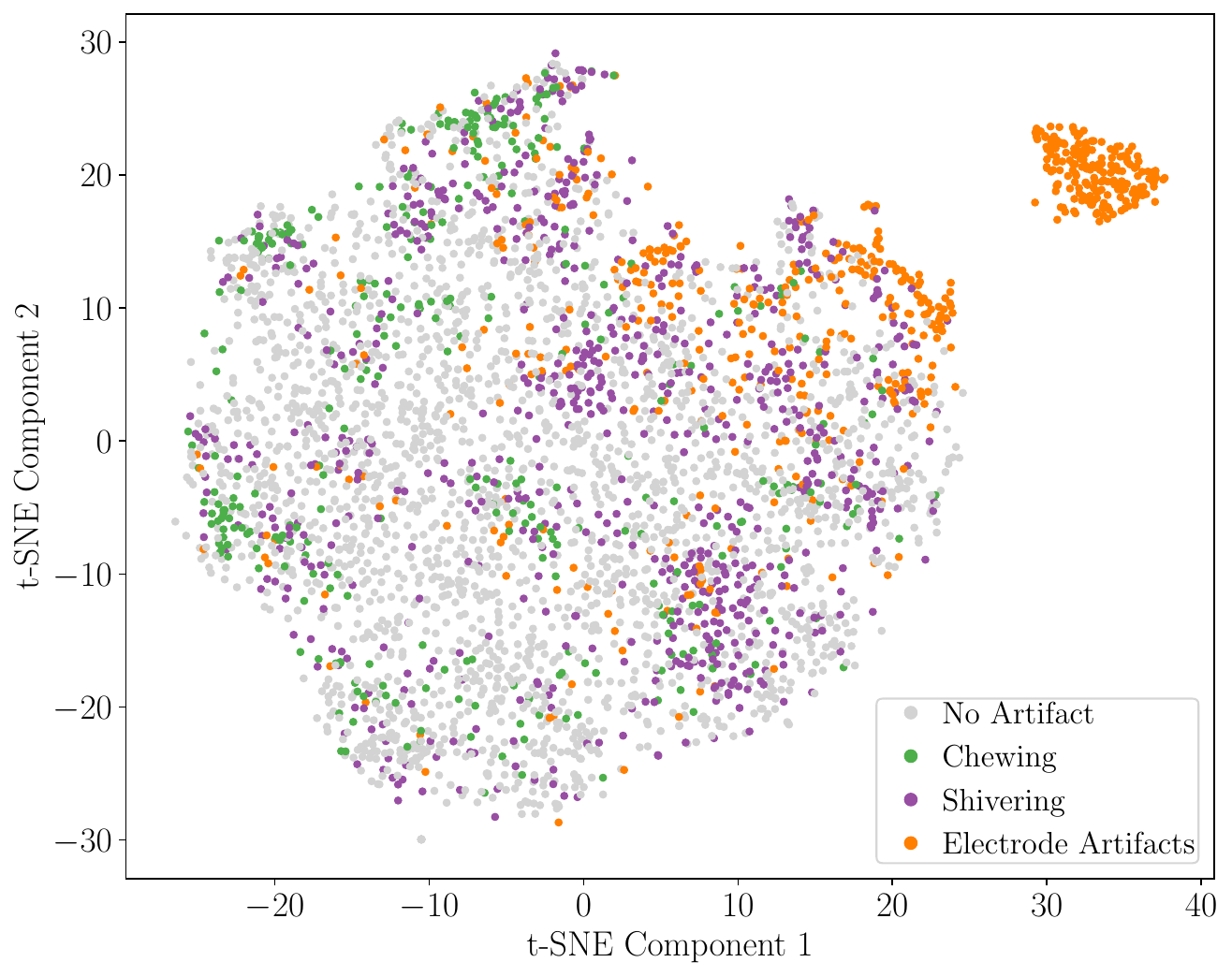} 
        \caption{TUAR dataset (Artifact Types).}
        \label{fig:app_TUAR_emb_raw}
    \end{subfigure}
    \caption{t-SNE of raw features on downstream datasets before fine-tuning.}
    \label{fig:t-sne-raw}
    \vspace{-0.5cm}
\end{figure}


\clearpage

\end{document}